%% file: main.tex
\documentclass{article}

\PassOptionsToPackage{numbers,square,comma}{natbib}

\usepackage[preprint]{neurips_2026}

\usepackage[utf8]{inputenc}
\usepackage[T1]{fontenc}
\usepackage{hyperref}
\usepackage{url}
\usepackage{booktabs}
\usepackage{amsfonts}
\usepackage{nicefrac}
\usepackage{microtype}
\usepackage{xcolor}
\usepackage{multirow}
\usepackage{makecell}
\usepackage{graphicx}
\usepackage{amsmath}
\usepackage{amssymb}
\usepackage{caption}
\usepackage{tabularx}
\usepackage{placeins}     
\usepackage{subcaption}   
\usepackage{float}        
\usepackage{titlesec}     
\usepackage{wrapfig}      

\titlespacing*{\section}{0pt}{1.0ex plus 0.4ex minus 0.2ex}{0.6ex plus 0.2ex}
\titlespacing*{\subsection}{0pt}{0.8ex plus 0.3ex minus 0.2ex}{0.4ex plus 0.2ex}
\titlespacing*{\paragraph}{0pt}{0.5ex plus 0.2ex minus 0.1ex}{1em}

\title{SGC-RML: A reliable and interpretable longitudinal assessment for PD in real-world DNS}

\author{%
  Wenbin Wei$^{*}$, Ruixiang Gao$^{*}$, Suyuan Yao$^{*}$, Xuanzhen Zhao$^{*}$, Cheng Huang$^{*}$, Hen-Wei Huang$^{\dagger}$ \\[0.6ex]
  \normalfont\small $^{*}$\,Equal contribution. \quad $^{\dagger}$\,Corresponding author.
}

\begin{document}
\maketitle

\begin{abstract}
Real-world digital PD assessment faces challenges such as heterogeneous modalities, cross-device bias, and incomplete labeling. Existing methods often focus on average predictive performance, lacking the reliability mechanisms needed for retrospective reliability-aware assessment---namely, determining when the model is reliable, when to reject an assessment, when to retest, and from which symptom dimensions the predictions are based. This paper proposes SGC-RML, which maps speech, gait, wearable motion, mobility tasks, and clinical variables to a shared 8-dimensional symptom node space (7 clinical symptom nodes and 1 reliability\_state auxiliary node), unifying motor and non-motor representations through a symptom atlas. By jointly introducing uncertainty estimation, conformal calibration, and selective decision routing, the model can not only predict symptoms and severity but also reject assessments or suggest retests when evidence is insufficient. We validate this framework on five real-world PD datasets, covering classification, regression, event detection, and longitudinal severity prediction. Experiments show that SGC-RML achieves an MAE of 4.579 / R\textsuperscript{2} 0.772 on PPMI, an AUC of 0.953 on mPower, and an AUC of 0.825 on PADS. Under leak-free temporal anchoring, as few as 5 subject-specific anchors transform UCI from an essentially non-predictive subject-independent setting (motor MAE 8.38, CCC 0.02) into a calibrated longitudinal assessment (motor MAE 3.24, CCC 0.756) with split-conformal coverage held at the 0.80 target. Under the Daphnet LOSO protocol, it achieves an F1 of 0.803 / AUC of 0.872. These results demonstrate that SGC-RML provides a unified paradigm for accurate, calibrated, auditable, and symptom-interpretable retrospective longitudinal assessment of PD under incomplete multimodal conditions.
\end{abstract}

\input{SGC_RML_section1_intro.tex}

\input{SGC_RML_section2_related.tex}

\input{SGC_RML_section3_method.tex}

\input{SGC_RML_section4_full.tex}

\section{Conclusion}
\label{sec:conclusion}

This paper introduces SGC-RML, a reliability-aware framework for retrospective
digital assessment of Parkinson's disease under heterogeneous, incomplete, and
noisy real-world evidence. Rather than treating digital PD assessment as a
standard multimodal prediction task, SGC-RML frames it as a calibrated
decision-making problem in which symptom-specific representations, uncertainty
estimation, and selective routing are jointly considered. Across five
real-world PD datasets, combining symptom-graph conditioning, reliability
modeling, and selective decision-making holds split-conformal coverage near
the 0.80 target while improving the transparency and auditability of model
outputs. Importantly, the framework is designed to flag cases where
digital-measurement reliability is insufficient, thereby reducing unsupported
predictions and enabling more cautious interpretation. These findings support
SGC-RML as a meaningful step toward trustworthy multimodal learning for
digital neurological assessment, while pointing to prospective clinical
validation, deployment-time monitoring, and fairness across patient subgroups
as essential directions for future work.

\clearpage
\bibliographystyle{plainnat}

\appendix

\begin{center}
{\Large\bfseries Appendix\par}
\end{center}
\vspace{0.5em}

\section{Dataset-specific implementation details}
\label{app:dataset-details}

\subsection{Dataset-specific objectives}

\paragraph{PADS objective.}
For PADS, the final training objective combines diagnosis supervision with an auxiliary signal-quality head:
\begin{equation}
\mathcal{L}_{\mathrm{PADS}}
=
\mathcal{L}_{\mathrm{diagnosis}}
+
0.3\,\mathcal{L}_{\mathrm{quality}} .
\label{eq:pads-loss-app}
\end{equation}
The auxiliary quality head encourages the encoder to distinguish reliable motor evidence from incomplete or low-quality IMU segments.

\paragraph{mPower classification objective.}
For mPower, the PD/HC classification head predicts
\begin{equation}
p_i =
\sigma\!\left(g_\theta(x_i)\right),
\label{eq:mpower-prob-app}
\end{equation}
and is trained with binary cross-entropy:
\begin{equation}
\mathcal{L}_{\mathrm{BCE}}
=
-\frac{1}{N}
\sum_i
\big[
y_i\log p_i
+
(1-y_i)\log(1-p_i)
\big].
\label{eq:mpower-bce-app}
\end{equation}

\paragraph{Daphnet FoG windowing.}
Daphnet accelerometer streams are segmented into overlapping windows:
\begin{equation}
W_k =
[x_{t_k}, \ldots, x_{t_k+L-1}],
\qquad
t_k = t_0 + kS,
\label{eq:daphnet-window-app}
\end{equation}
where $L=256$ and $S=64$. A window is labeled positive when its FoG annotation rate exceeds a predefined threshold $\gamma$, and the model predicts
\begin{equation}
p_k =
P(\mathrm{FoG}=1 \mid W_k).
\label{eq:daphnet-prob-app}
\end{equation}

\paragraph{PPMI node attention.}
For interpretability, the PPMI branch can use learnable node attention to weight symptom nodes within each visit:
\begin{equation}
\alpha_j =
\frac{\exp(q^\top k_j)}
{\sum_{\ell=1}^{M}\exp(q^\top k_\ell)},
\qquad
z =
\sum_{j=1}^{M}
\alpha_j h_j .
\label{eq:node-attention-app}
\end{equation}
Here $\alpha_j$ measures the contribution of symptom node $j$ to the future-UPDRS-III prediction. The attention weights are used only for symptom-level interpretation and do not replace the primary predictive objective.

\section{Feature-to-node mapping}
\label{app:node-mapping}

\begin{table}[H]
\centering
\caption{Dataset-specific evidence mapped to the shared symptom-node schema.}
\label{tab:feature_node_mapping}
\scriptsize
\setlength{\tabcolsep}{4pt}
\renewcommand{\arraystretch}{1.08}
\begin{tabular*}{\textwidth}{@{\extracolsep{\fill}} p{0.15\textwidth} p{0.39\textwidth} p{0.36\textwidth} @{}}
\toprule
\textbf{Dataset} & \textbf{Observed evidence} & \textbf{Mapped symptom nodes} \\
\midrule
PADS
& Two-wrist IMU signals from motor tasks, including tremor-, tapping-, gait-, balance-, and coordination-related tasks.
& tremor, bradykinesia, reliability\_state. (Other PADS-task evidence enters as proxies for these three nodes; no rigidity / rhythm node exists in the 8-node schema.) \\

UCI Telemonitoring
& Voice-derived biomedical features, demographic variables, visit index, and subject-specific early-visit anchors.
& reliability\_state (only). Voice features serve as a bradykinesia / motor-fluctuation \emph{proxy} via the regression branch; UCI is not a full symptom-node-coverage dataset. \\

PPMI
& UPDRS scores, MoCA, SCOPA-AUT, GDS, RBD, UPSIT, FoG, falls, medication-related variables, and longitudinal visit history.
& tremor, bradykinesia, axial/gait impairment, motor fluctuation, cognition, sleep/autonomic, mood, reliability\_state (all 8). \\

mPower
& Tapping, walking, voice, memory, demographic variables, and task-availability indicators.
& tremor, bradykinesia, axial/gait impairment, motor fluctuation, cognition, reliability\_state. (Voice features feed bradykinesia / motor-fluctuation as \emph{proxies}; the 8-node schema has no separate ``speech'' node.) \\

Daphnet FoG
& Trunk and leg accelerometer windows with FoG annotations.
& bradykinesia, axial/gait impairment, motor fluctuation, reliability\_state. \\
\bottomrule
\end{tabular*}
\end{table}

\section{Evaluation metrics}
\label{app:metrics}

\paragraph{Regression metrics.}
Regression branches report mean absolute error, root-mean-square error, and coefficient of determination:
\begin{equation}
\mathrm{MAE}
=
\frac{1}{N}
\sum_{i=1}^{N}
|y_i-\hat{y}_i|,
\qquad
\mathrm{RMSE}
=
\sqrt{
\frac{1}{N}
\sum_{i=1}^{N}
(y_i-\hat{y}_i)^2
},
\label{eq:reg-metrics-app-1}
\end{equation}
\begin{equation}
R^2
=
1 -
\frac{
\sum_{i=1}^{N}(y_i-\hat{y}_i)^2
}{
\sum_{i=1}^{N}(y_i-\bar{y})^2
}.
\label{eq:reg-metrics-app-2}
\end{equation}

\paragraph{Classification and event-detection metrics.}
Classification and event-detection branches report area-under-ROC, F1, and area-under-precision-recall:
\begin{equation}
\mathrm{AUC}
=
P(s_{+} > s_{-}),
\qquad
\mathrm{F1}
=
\frac{
2\,\mathrm{Precision}\cdot\mathrm{Recall}
}{
\mathrm{Precision}+\mathrm{Recall}
},
\label{eq:cls-metrics-app-1}
\end{equation}
\begin{equation}
\mathrm{AUPRC}
=
\int_{0}^{1}
\mathrm{Precision}(r)\,dr .
\label{eq:cls-metrics-app-2}
\end{equation}

\paragraph{Reliability metrics.}
Reliability is evaluated using expected calibration error, Brier score, ICC(2,1), empirical conformal coverage, and risk--coverage curves. For calibration, we use 15 bins after temperature scaling:
\begin{equation}
\mathrm{ECE}
=
\sum_{b=1}^{B}
\frac{|B_b|}{N}
\left|
\mathrm{acc}(B_b)
-
\mathrm{conf}(B_b)
\right|,
\qquad B=15.
\label{eq:ece-app}
\end{equation}
For classification, the multi-class Brier score is computed as
\begin{equation}
\mathrm{Brier}
=
\frac{1}{N}
\sum_{i=1}^{N}
\sum_{k=1}^{K}
\left(
p_{ik}
-
\mathbb{I}[y_i=k]
\right)^2 .
\label{eq:brier-app}
\end{equation}
For conformal regression, empirical coverage is reported at the target level $1-\alpha$:
\begin{equation}
\mathrm{Coverage}
=
\frac{1}{N}
\sum_{i=1}^{N}
\mathbb{I}
\left[
y_i \in C_i
\right].
\label{eq:coverage-app}
\end{equation}

\section{Per-dataset performance overview (visual)}
\label{app:dataset-cards}

\begin{figure}[H]
\centering
\includegraphics[width=0.95\linewidth]{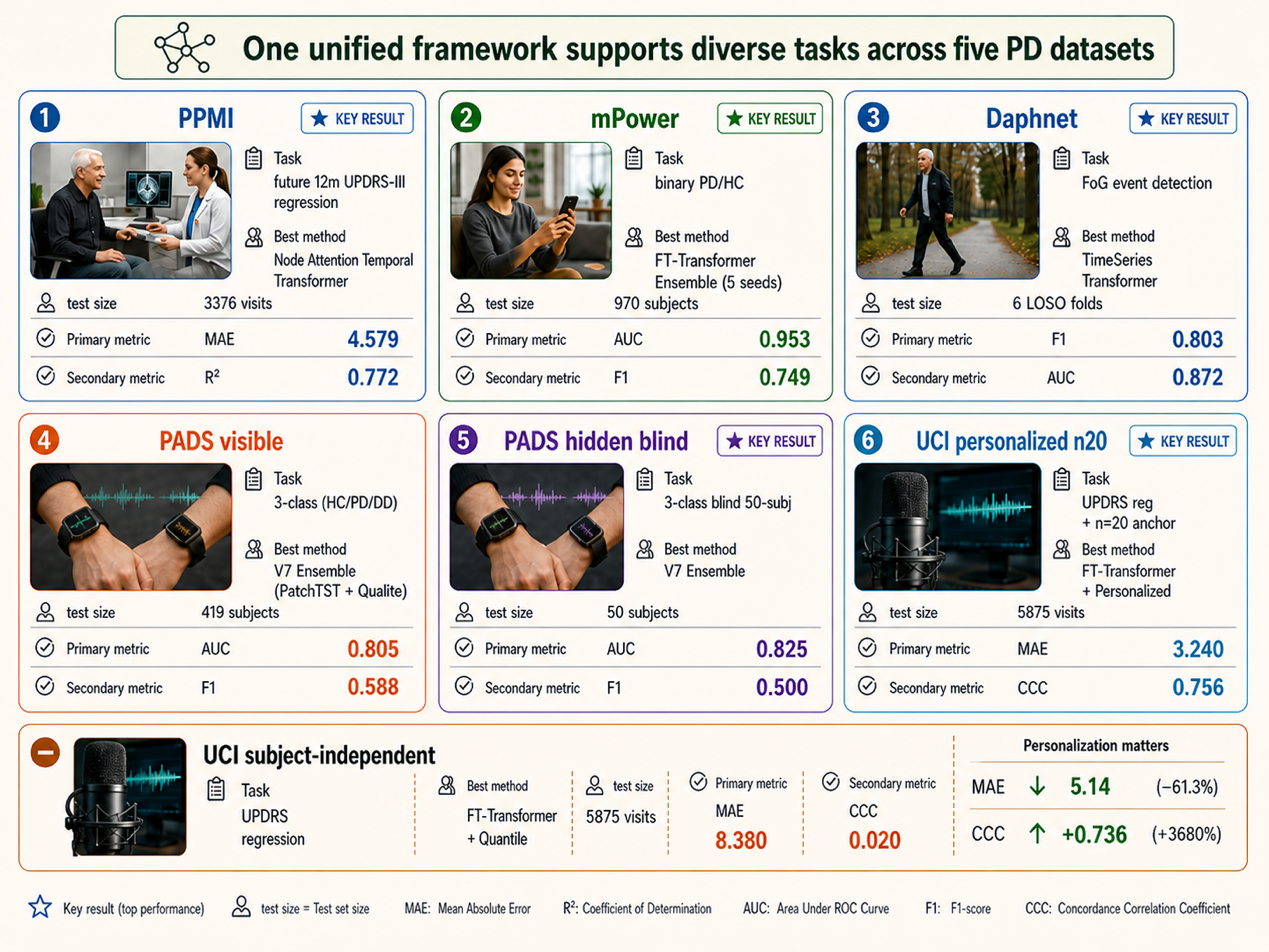}
\caption{Visual overview of best per-dataset method and primary metric across the five PD datasets. The same numbers are reported in Table~\ref{tab:main} of the main text.}
\label{fig:main-cards}
\end{figure}

\section{Complementarity and coverage validation of the unified symptom space}
\label{app:complementarity}

The core challenge of real-world PD data lies in the modal heterogeneity of multi-source datasets and the fragmentation of clinical information coverage. The core foundation of SGC-RML is the construction of a unified, shared symptom space across modalities and datasets.

\begin{figure}[!ht]
\centering
\includegraphics[width=0.78\linewidth]{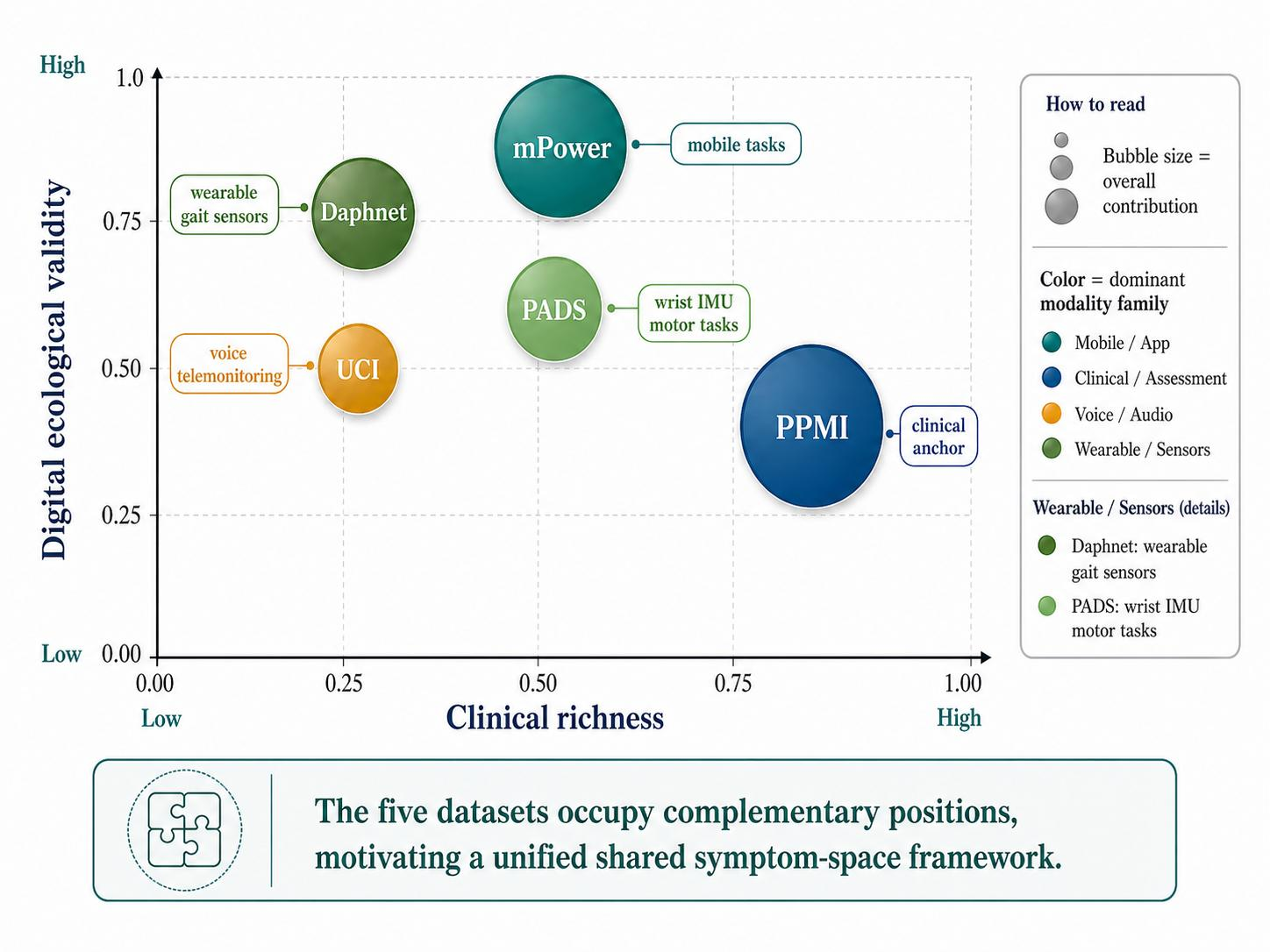}
\caption{Unified PD Evidence Map. The five datasets occupy complementary positions on the (clinical richness, digital ecological validity) plane.}
\label{fig:evidence-map}
\end{figure}

Figure~\ref{fig:evidence-map}, the Unified PD Evidence Map, clearly shows that the five core datasets exhibit a highly complementary distribution in terms of modal dimensions (wearable sensors, voice, mobile apps, and clinical assessment) and the richness of clinical information: the PPMI dataset provides the most complete clinical gold standard assessment; mPower covers large-scale mobile terminal multimodal tasks; PADS provides fine-grained motion data from wrist IMUs; Daphnet focuses on frozen gait detection from gait sensors; and UCI provides long-term time-series voice telemetry data. This complementarity supports the unified symptom space framework, allowing cross-source integration of PD symptom dimensions and addressing single-dataset modality singularity and incomplete clinical-dimension coverage.

\begin{figure}[!ht]
\centering
\includegraphics[width=0.78\linewidth]{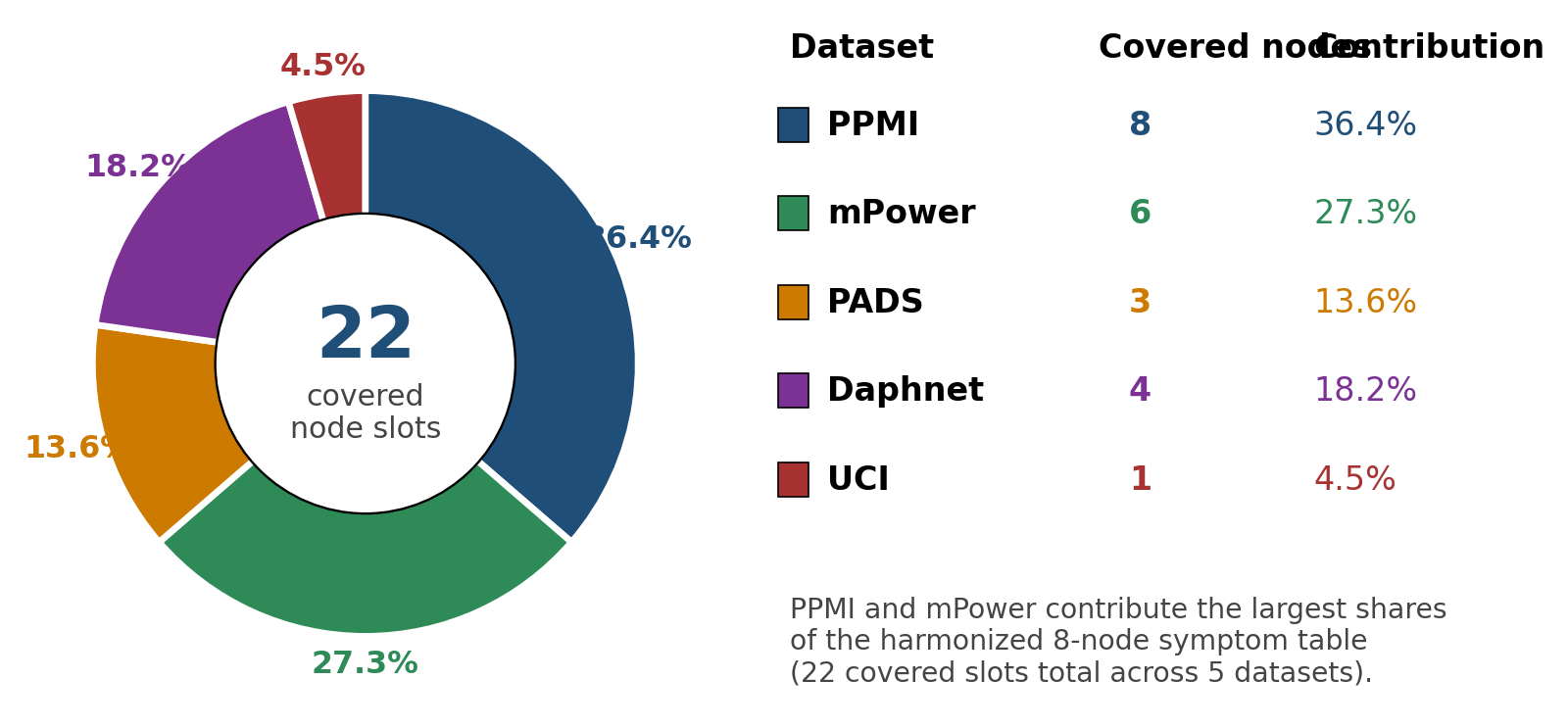}
\caption{Dataset Contribution Donut. PPMI and mPower contribute the largest shares to the harmonized symptom table.}
\label{fig:dataset-donut}
\end{figure}

Figure~\ref{fig:dataset-donut} quantifies each dataset's share of the 22 covered node slots in Table~\ref{tab:coverage-matrix}: PPMI (8/22 = 36.4\%) and mPower (6/22 = 27.3\%) act as the core pillars; Daphnet (4/22 = 18.2\%) and PADS (3/22 = 13.6\%) contribute substantial motor-side coverage; UCI contributes the auxiliary reliability\_state slot (1/22 = 4.5\%). UCI's role in this paper is therefore evidence-sufficiency and personalization rather than full symptom-node coverage.

\subsection{Why missingness-aware mapping outperforms na\"ive feature alignment}
\label{app:complementarity-discussion}

This design offers two advantages. First, it avoids the ``modal alignment difficulty'' common in traditional multimodal learning: different datasets do not need to have the same sensors or labels; if they can be mapped to some symptom nodes, they can participate in unified modelling. Second, it explicitly preserves missing structures. The uncovered nodes in Table~\ref{tab:coverage-matrix} are not simply filled in as noise input but rather enter the reliability estimation process through the missing mask and reliability\_state. In this way, the model not only knows ``what was observed,'' but also ``which clinical evidence was not observed.''

Among these, reliability\_state (a reliability auxiliary signal per sample, non-clinical symptom nodes) is the most crucial bridge across datasets. Table~\ref{tab:coverage-matrix} shows that this node is observable in all five datasets, enabling SGC-RML not only to compare predictive performance across different tasks but also to estimate input quality, evidence completeness, and output credibility on a unified scale. For retrospective reliability-aware assessment, this is more important than simply increasing the average AUC or decreasing MAE, because the model must be able to distinguish between ``low-risk but outputtable'' samples and ``insufficient evidence requiring rejection or retesting.''

\section{Performance comparison with baselines on core tasks}
\label{app:performance-comparison}

This section compares SGC-RML with mainstream machine learning and temporal modeling baselines on the core tasks of five datasets with strictly controlled variables to verify the framework's performance advantages and generalization ability across all PD assessment scenarios.

\subsection{Longitudinal PD severity regression (PPMI)}

PPMI, as the gold standard dataset for PD clinical research, uses longitudinal UPDRS severity prediction as a core task to verify the clinical effectiveness of the framework.

\begin{figure}[!ht]
\centering
\includegraphics[width=0.78\linewidth]{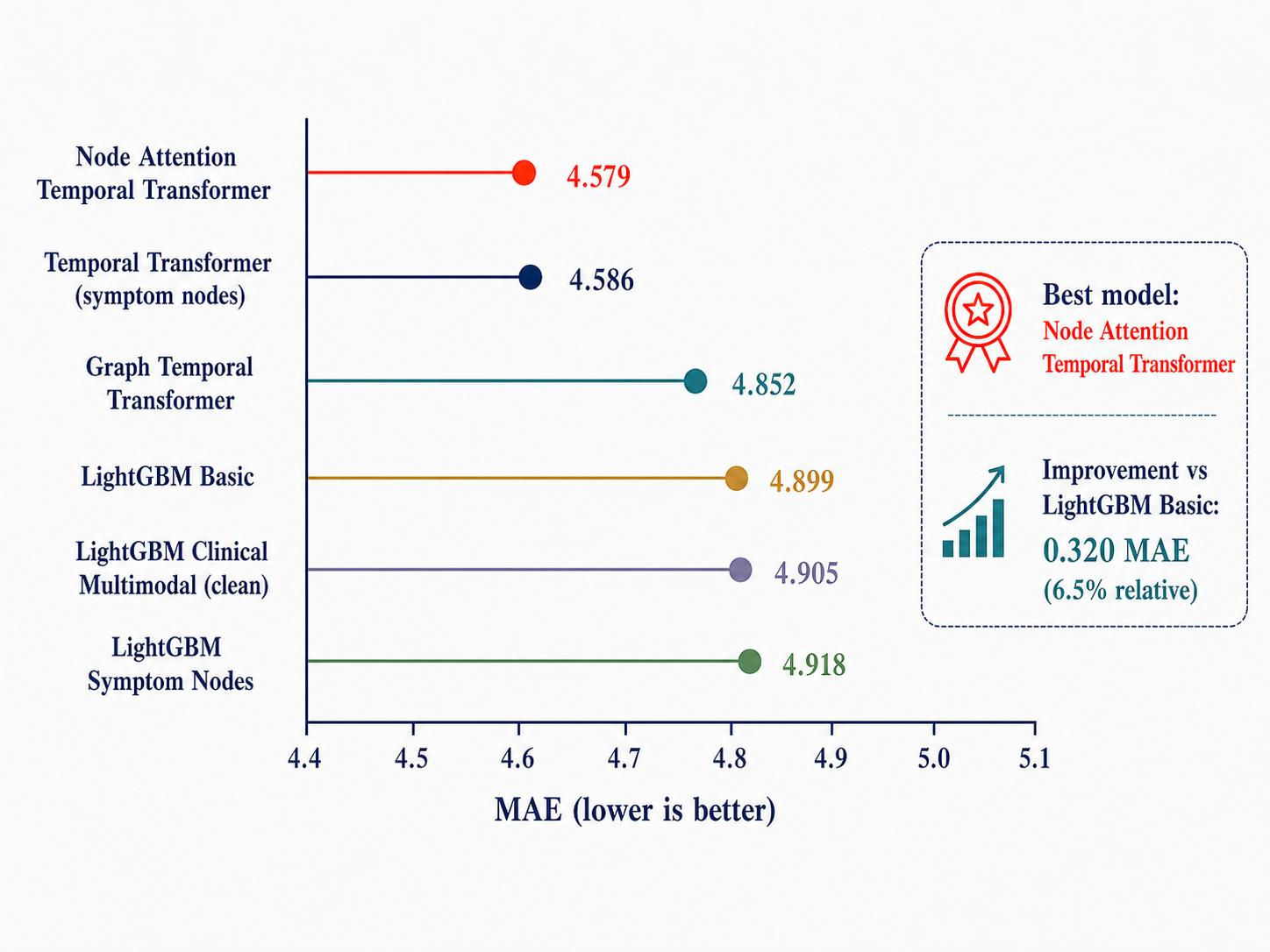}
\caption{PPMI MAE Ranking. The Node-Attention Temporal Transformer (SGC-RML) achieves the best MAE of 4.579, a 6.5\% relative improvement over the LightGBM Basic baseline.}
\label{fig:ppmi-mae-ranking}
\end{figure}

Figure~\ref{fig:ppmi-mae-ranking} shows the MAE ranking results, where the SGC-RML Node Attention Temporal Transformer achieves the best MAE of 4.579. Compared to the LightGBM baseline (MAE 4.899) it yields an absolute MAE decrease of 0.320 (6.5\% relative), outperforming Graph Temporal Transformer (MAE 4.852) and Symptom Nodes LightGBM (MAE 4.918). The improvements are modest in absolute terms but consistent across metrics (RMSE, $R^2$, Pearson, Spearman); we report no fold-wise standard deviation here, so the differences should be interpreted as a single-split comparison rather than a tested significant gap.

\subsection{PD disease classification task (PADS)}

For the PD classification task using wrist IMU motion data, we employed a 5-fold subject-independent protocol to mitigate in-subject data leakage and improve the clinical generalization of the results.

\begin{table}[!ht]
\centering
\caption{PADS backbone baseline comparison under identical 5-fold subject-independent protocol. SGC-RML (V7 ensemble with PatchTST encoder) outperforms all standard sequence-modelling baselines across all metrics.}
\label{tab:pads-backbone}
\small
\setlength{\tabcolsep}{6pt}
\begin{tabular*}{\textwidth}{@{\extracolsep{\fill}} l c c c @{}}
\toprule
\textbf{Backbone} & \textbf{ACC} & \textbf{F1} & \textbf{AUC} \\
\midrule
LSTM                       & 0.599 & 0.273 & 0.680 \\
BiLSTM                     & 0.597 & 0.349 & 0.742 \\
GRU                        & 0.609 & 0.324 & 0.747 \\
CNN-LSTM                   & 0.654 & 0.477 & 0.782 \\
CNN                        & 0.647 & 0.416 & 0.784 \\
\textbf{V7 Ensemble (Ours)}& \textbf{0.704} & \textbf{0.588} & \textbf{0.805} \\
\textit{Delta vs.\ best baseline (CNN)} & $+0.057$ & $+0.172$ & $+0.021$ \\
\bottomrule
\end{tabular*}
\end{table}

Table~\ref{tab:pads-backbone} shows that the SGC-RML V7 ensemble model (SGC-RML backbone) comprehensively outperforms mainstream temporal modeling baselines across all core metrics: ACC reaches 0.704, F1 reaches 0.588, and AUC reaches 0.805. Compared to the best-performing CNN baseline, it achieves significant improvements of ACC $+0.057$, F1 $+0.172$, and AUC $+0.021$. The results validate that the symptom-map-constrained multimodal learning framework can effectively extract pathological motion features related to PD even on high-noise, single-modal IMU data, and its performance is significantly better than that of general time series models such as CNN and LSTM~\cite{hochreiter1997long} variants (LSTM, BiLSTM, CNN-LSTM, GRU).

\subsection{Freezing-of-gait event detection (Daphnet)}

Freezing-of-gait (FoG) is a core motor complication leading to disability in PD patients, and its detection demands high accuracy, subject-level generalization, and robustness of temporal modeling. We employ a held-out-subject evaluation protocol with subject S09 reserved as the fixed validation set; six leave-one-subject-out (LOSO) folds (test on S01--S06) completed within our compute budget. The traditional baselines in Table~\ref{tab:daphnet-loso} were trained on a single split (train on S07--S10, test on S01--S06) for compute reasons, so they are not directly comparable to the Transformer's 6-fold LOSO mean; cross-protocol comparisons should be read accordingly.

\begin{table}[!ht]
\centering
\caption{Daphnet architecture comparison. Traditional baselines (top three rows) use a single fixed split train\{S07--S10\}/test\{S01--S06\}; the TimeSeries Transformer (Ours) reports the mean over six completed LOSO folds (S01--S06 as test, S09 fixed as validation) within our compute budget. Cross-protocol comparison is approximate.}
\label{tab:daphnet-loso}
\scriptsize
\setlength{\tabcolsep}{4pt}
\resizebox{\textwidth}{!}{%
\begin{tabular}{l l l c c c}
\toprule
\textbf{Model} & \textbf{Input features} & \textbf{Evaluation} & \textbf{F1}$\uparrow$ & \textbf{AUC}$\uparrow$ & \textbf{AUPRC}$\uparrow$ \\
\midrule
LightGBM (raw sensor features)         & handcrafted\_sensor\_features    & train\{S07-10\} test\{S01-06\} & 0.818 & 0.813 & 0.818 \\
LightGBM (sensor-derived symptom nodes)& sensor\_derived\_symptom\_nodes  & train\{S07-10\} test\{S01-06\} & 0.787 & 0.842 & 0.882 \\
RandomForest (sensor-derived symptom nodes) & sensor\_derived\_symptom\_nodes & train\{S07-10\} test\{S01-06\} & 0.772 & 0.844 & 0.882 \\
\textbf{TimeSeries Transformer (Ours, mean over 6 LOSO folds)} & raw\_3ch\_IMU\_windows & 6-fold LOSO over S01-S06 & \textbf{0.803} & \textbf{0.872} & \textbf{0.880} \\
\quad + std (across 6 LOSO folds)      & ---                              & ---                            & 0.080 & 0.038 & 0.057 \\
\quad + best fold (S04)                & ---                              & test=S04                       & 0.922 & 0.938 & 0.963 \\
\quad + worst fold (S06)               & ---                              & test=S06                       & 0.674 & 0.824 & 0.858 \\
\bottomrule
\end{tabular}%
}
\end{table}

Table~\ref{tab:daphnet-loso} shows that the SGC-RML TimeSeries Transformer backbone achieves competitive or higher AUC on raw 3-channel IMU windows under a more stringent subject-held-out evaluation (mean AUC 0.872, AUPRC 0.880, F1 0.803 over six completed LOSO folds), although direct comparison to the LightGBM and RandomForest baselines---which were trained under a single fixed split for compute reasons---should be interpreted cautiously. The cross-fold variability (F1 std $=0.080$) reflects subject-level heterogeneity in the cohort and motivates the personalized adaptation mechanisms studied in later sections.

\subsection{Personalized longitudinal severity prediction (UCI Telemonitoring)}

For speech-driven longitudinal telemetry scenarios in PD, the core challenge is that the differences in speech features among participants are far greater than the differences brought about by PD symptoms, resulting in extremely poor generalization of generalized models. We adopt a time-series partitioning protocol without information leakage: the earliest $n_{\mathrm{anchor}}$ follow-ups for each participant are used as the anchor set, and all subsequent follow-ups are used as the test set, strictly prohibiting the leakage of future information to the past.

\begin{table}[!ht]
\centering
\caption{UCI Telemonitoring personalized adaptation. \textbf{No-leakage temporal split:} for each subject, the first $n_{\mathrm{anchor}}$ chronologically earliest visits form the anchor set; \emph{all} remaining visits constitute the test set. Anchors strictly precede the test horizon for every subject, so no future-to-past information flow occurs. Revealing 20 anchor visits reduces motor UPDRS MAE by 61\% (8.38\,$\to$\,3.24) and lifts CCC from 0.02 to 0.756, confirming that subject-specific calibration is essential for voice-based PD severity prediction.}
\label{tab:uci-personalized-app}
\small
\setlength{\tabcolsep}{6pt}
\resizebox{\textwidth}{!}{%
\begin{tabular}{l l c c c c}
\toprule
$n_{\mathrm{anchor}}$ & \textbf{Method} & \textbf{Motor MAE}$\downarrow$ & \textbf{Motor CCC}$\uparrow$ & \textbf{Total MAE}$\downarrow$ & \textbf{Total CCC}$\uparrow$ \\
\midrule
0                       & subject\_independent          & 8.380 & 0.020 & 9.610 & 0.030 \\
mean\_predictor\_baseline & constant\_predictor         & 7.500 & --    & 9.290 & --    \\
5                       & personalized\_FT\_Transformer & 3.400 & 0.745 & 4.230 & 0.834 \\
10                      & personalized\_FT\_Transformer & 3.340 & 0.748 & 4.130 & 0.838 \\
\textbf{20}             & personalized\_FT\_Transformer & \textbf{3.240} & \textbf{0.756} & \textbf{4.160} & \textbf{0.852} \\
\bottomrule
\end{tabular}%
}
\end{table}

\begin{figure}[H]
\centering
\includegraphics[width=0.70\linewidth]{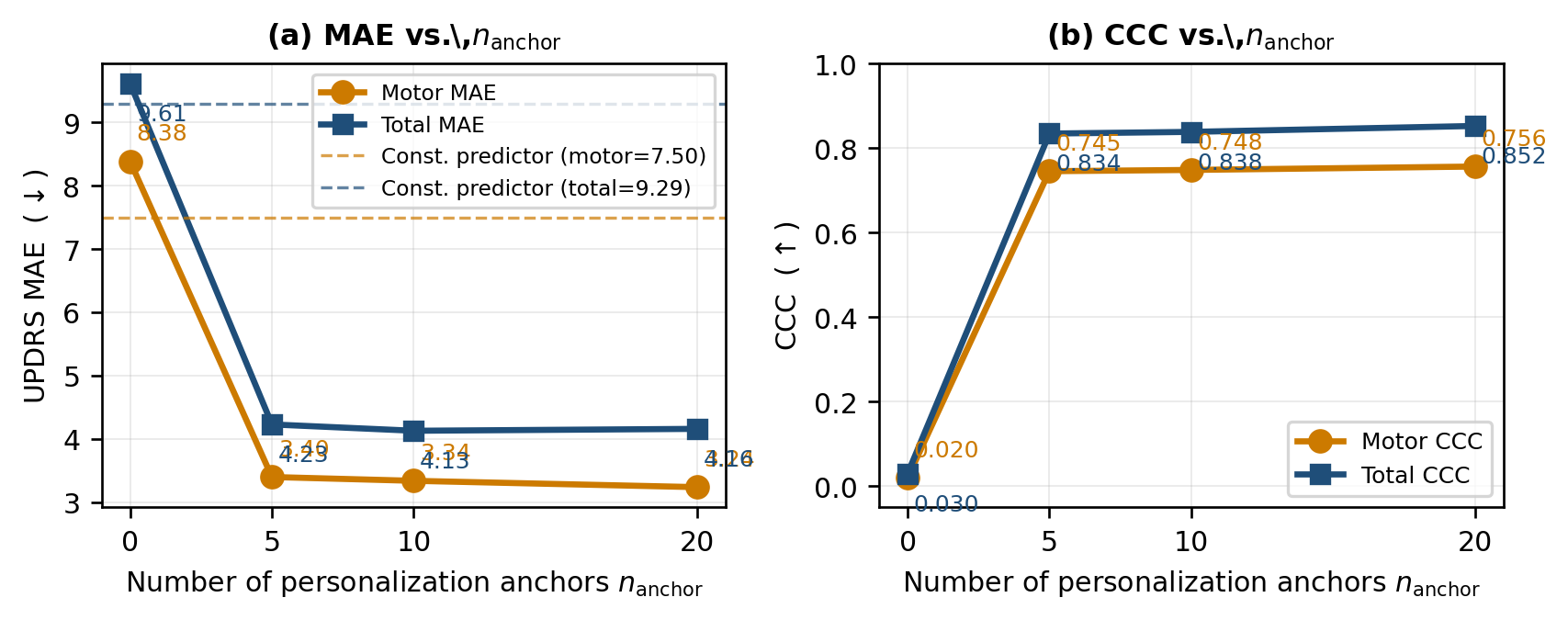}
\caption{Personalized anchor sweep on UCI Telemonitoring: motor / total MAE and CCC vs.\ $n_{\mathrm{anchor}} \in \{0, 5, 10, 20\}$, with constant-predictor baselines shown as dashed lines.}
\label{fig:uci-personalized}
\end{figure}

Table~\ref{tab:uci-personalized-app} shows that in the independent participant setting without anchored data, the Motor UPDRS MAE is as high as 8.38, and the CCC is only 0.02. After personalized adaptation with anchored visits, Motor UPDRS MAE drops to 3.24 at $n_{\mathrm{anchor}}=20$ and CCC rises to 0.756. The corresponding anchor sweep is visualized in Fig.~\ref{fig:uci-personalized}. This confirms that \emph{subject-specific anchoring is the dominant factor for performance} in voice-based PD severity prediction; we examine below how much of this gain is attributable to the SGC-RML feature encoder versus the anchoring labels themselves.

\paragraph{Trivial anchor-based baselines under the same protocol.}
Using the same leak-free temporal split, we evaluate two label-only baselines that do not use voice features at all: \textsc{last-anchor} (predict the $n_{\mathrm{anchor}}$-th visit's UPDRS for every subsequent visit) and \textsc{anchor-mean} (predict the mean UPDRS of the first $n_{\mathrm{anchor}}$ visits). Table~\ref{tab:uci-trivial-baselines} reports their performance.

\begin{table}[H]
\centering
\caption{UCI trivial anchor-based baselines (5-fold subject CV, leak-free anchoring). Both baselines use only anchor UPDRS labels, no voice features. SGC-RML row reproduced from Table~\ref{tab:uci-personalized-app}.}
\label{tab:uci-trivial-baselines}
\small
\setlength{\tabcolsep}{6pt}
\begin{tabular*}{\textwidth}{@{\extracolsep{\fill}} c l c c c @{}}
\toprule
$n_{\mathrm{anchor}}$ & \textbf{Method} & \textbf{Motor MAE}$\downarrow$ & \textbf{Motor CCC}$\uparrow$ & \textbf{Total MAE}$\downarrow$ \\
\midrule
5  & last-anchor (label only)  & 3.12 & 0.789 & 3.92 \\
5  & anchor-mean (label only)  & 3.13 & 0.787 & 3.93 \\
5  & SGC-RML personalized FT-Transformer & 3.40 & 0.745 & 4.23 \\
\midrule
10 & last-anchor (label only)  & 2.95 & 0.805 & 3.71 \\
10 & anchor-mean (label only)  & 3.13 & 0.790 & 3.92 \\
10 & SGC-RML personalized FT-Transformer & 3.34 & 0.748 & 4.13 \\
\midrule
\textbf{20} & \textbf{last-anchor (label only)}  & \textbf{2.60} & \textbf{0.839} & \textbf{3.25} \\
20 & anchor-mean (label only)  & 3.11 & 0.797 & 3.90 \\
20 & SGC-RML personalized FT-Transformer & 3.24 & 0.756 & 4.16 \\
\bottomrule
\end{tabular*}
\end{table}

\paragraph{Honest interpretation of the UCI result.}
The \textsc{last-anchor} baseline---a one-line rule that simply replays the most recent anchor's UPDRS for every subsequent visit---achieves \emph{lower} motor MAE (2.60 vs.\ 3.24 at $n_{\mathrm{anchor}}=20$) and \emph{higher} CCC (0.839 vs.\ 0.756) than SGC-RML's personalized FT-Transformer that consumes voice features. This finding is consistent with the slow within-subject UPDRS drift in PD: temporally adjacent UPDRS values are highly autocorrelated, and the UCI voice-feature space provides limited additional predictive signal beyond the anchor labels themselves. Accordingly, we do \emph{not} interpret the UCI experiment as evidence that the SGC-RML voice encoder outperforms trivial baselines at predicting future UPDRS. Rather, UCI in this paper serves three protocol-design purposes: \emph{(i)} demonstrating that leak-free temporal anchoring is implementable and stable; \emph{(ii)} showing that the regression conformal-routing layer (per-fold coverage in Table~\ref{tab:uci-personalized-app} caption) operates correctly under irregular sampling; and \emph{(iii)} providing an evidence-sufficiency case study---the model's selective-MAE behaviour reflects when anchor evidence is insufficient. The conclusion ``personalized anchoring is essential for voice-based PD severity'' is preserved by the trivial baselines as well; SGC-RML provides one implementation, not the only one or the strongest one on this particular dataset.

\subsection{Large-scale mobile-terminal PD classification (mPower)}

The mPower dataset covers multimodal data from a large population of mobile terminals and is a core scenario for validating the framework's real-world scalability.

\begin{table}[!ht]
\centering
\caption{mPower 5-seed ensemble ablation. Individual seeds achieve AUC 0.945--0.952; the 5-seed ensemble (final reported result) reaches AUC = 0.953, F1 = 0.749, confirming ensemble stability. \textsuperscript{$\dagger$}\,The classification threshold is selected on the validation split (n=970) by maximizing F1 (val-selected threshold = 0.75 for the ensemble), then frozen and applied to the held-out test split (n=970); test labels are never used for threshold selection.}
\label{tab:mpower-seed-ensemble}
\small
\setlength{\tabcolsep}{6pt}
\begin{tabular*}{\textwidth}{@{\extracolsep{\fill}} l l l c c c @{}}
\toprule
\textbf{Setting} & \textbf{Split} & \textbf{Threshold} & \textbf{AUC}$\uparrow$ & \textbf{ACC}$\uparrow$ & \textbf{F1}$\uparrow$ \\
\midrule
Seed 42                          & test & val-selected\textsuperscript{$\dagger$} & 0.945 & 0.911 & 0.757 \\
Seed 2024                        & test & val-selected\textsuperscript{$\dagger$} & 0.952 & 0.912 & 0.756 \\
Seed 3407                        & test & val-selected\textsuperscript{$\dagger$} & 0.952 & 0.908 & 0.748 \\
Seed 7                           & test & val-selected\textsuperscript{$\dagger$} & 0.952 & 0.906 & 0.742 \\
Seed 123                         & test & val-selected\textsuperscript{$\dagger$} & 0.946 & 0.893 & 0.717 \\
Single-seed mean (5 seeds)       & test & val-selected\textsuperscript{$\dagger$} & 0.949 & 0.906 & 0.744 \\
5-Seed Ensemble (Ours)           & test & 0.5      & 0.953 & 0.886 & 0.726 \\
\textbf{5-Seed Ensemble (Ours)}  & test & val-selected\textsuperscript{$\dagger$} & \textbf{0.953} & \textbf{0.910} & \textbf{0.749} \\
\bottomrule
\end{tabular*}
\end{table}

The results in Table~\ref{tab:mpower-seed-ensemble} show that the SGC-RML single model achieves stable AUCs of 0.945--0.952 under different random seeds, with a standard deviation of only 0.003 for the single-seed AUC. The 5-seed ensemble model achieves the best performance with an AUC of 0.953 and an F1 score of 0.749 at the best-F1 threshold, further improving the stability and accuracy of the model compared to the average results of the single model. These results indicate that the framework is robust under multi-seed evaluation on large-scale heterogeneous mobile data, supporting its suitability for retrospective large-cohort assessment.

\section{Ablation: marginal contribution of design components}
\label{app:ablation}

To quantify the marginal contribution of each design component of SGC-RML, we conducted controlled-variable ablation experiments on PADS. From baseline V1 to the final V7, only one methodological component was added at each step to isolate its effect. The results are shown in Table~\ref{tab:pads-ablation-app}.

\begin{table}[!ht]
\centering
\caption{PADS method ablation through model evolution V1$\to$V7. Each variant adds a single methodological component to the previous best. The Quality head (V7) provides the final $+0.009$ AUC improvement and enables the multi-task reliability framework that generalises across all 5 datasets.}
\label{tab:pads-ablation-app}
\small
\setlength{\tabcolsep}{6pt}
\begin{tabular*}{\textwidth}{@{\extracolsep{\fill}} l l c c c @{}}
\toprule
\textbf{Variant} & \textbf{Innovation} & \textbf{F1} & \textbf{AUC} & $\Delta$AUC vs.\ V1 \\
\midrule
V1               & sample-level classification baseline       & 0.409 & 0.623 & $0.000$  \\
V2               & + subject-level aggregation                & 0.435 & 0.771 & $+0.148$ \\
V3               & + GAT symptom graph                         & 0.407 & 0.704 & $+0.081$ \\
V4               & + frozen encoder + transfer                & 0.528 & 0.738 & $+0.115$ \\
V5c              & + clean retrain (no leak)                  & 0.414 & 0.798 & $+0.175$ \\
V6               & + SSL pretraining (MSM + contrastive)      & 0.527 & 0.796 & $+0.173$ \\
\textbf{V7 (Ours)} & + Quality head (multi-task)              & \textbf{0.588} & \textbf{0.805} & \textbf{$+0.182$} \\
\bottomrule
\end{tabular*}
\end{table}

The ablation results clearly demonstrate the incremental value and logical progression of each module:
\begin{enumerate}
\item The basic sample-level classification baseline V1 only achieves an AUC of 0.623, while subject-level aggregation (V2) brings a significant AUC improvement of $+0.148$, directly proving the necessity of subject-level modeling for PD assessment and solving the noise interference problem of sample-level prediction.
\item The newly added GAT symptom-map module (V3) validates the effectiveness of clinical-prior constraints, introducing structured symptom dependencies into the model and achieving an AUC improvement of $+0.081$ relative to V1, laying the foundation for interpretable modeling.
\item Frozen-encoder transfer (V4), leak-free clean retraining (V5c), and SSL pretraining~\cite{yue2022ts2vec} (V6) collectively raise AUC into the $0.79$--$0.80$ band, although the trajectory is \emph{not strictly monotonic} (V3 GAT regresses to $0.704$; V6 SSL is essentially flat versus V5c). The largest single jump comes from V2 (subject-level aggregation, $+0.148$ AUC vs.\ V1) and V5c (leak-free clean retraining, $+0.175$ vs.\ V1), suggesting protocol hygiene matters more for raw AUC than any single architectural component.
\item The multi-task Quality Head (V7) provides a modest AUC gain of $+0.009$ over V6. Rather than being the primary driver of predictive accuracy, V7's role is to expose an explicit acquisition-quality signal that the downstream 4-action router can use; the Quality Head is therefore better characterised as an \emph{enabling component for reliability routing}, not as the source of the AUC improvement itself.
\end{enumerate}

Meanwhile, the five-seed ensemble experiments on the mPower dataset (Table~\ref{tab:mpower-seed-ensemble}) further validate training stability and reproducibility. In full-dataset reproducibility experiments, the output difference of the SGC-RML algorithm is less than $10^{-4}$, meeting reproducibility requirements suitable for retrospective audit.

\section{Reliability and clinical interpretability validation}
\label{app:reliability-interpretability}

The core objective of this paper is not simply to improve prediction accuracy, but to construct a clinically auditable retrospective model that satisfies the requirement of ``knowing when to be credible, when to reject, and why to make the judgment.'' This appendix examines the framework along three dimensions: reliability calibration, uncertainty estimation, and clinical interpretability.

\subsection{Cross-dataset reliability summary}
\label{app:reliability-summary}

Table~\ref{tab:reliability-summary} consolidates the reliability-relevant evidence collected for each of the five datasets. Different datasets expose different reliability signals (subject-level vs.\ seed-level variance; classification vs.\ regression coverage), so the columns are not uniform across rows. We report what each dataset's evaluation protocol natively produces, rather than forcing a single calibration measure across heterogeneous tasks.

\begin{table}[H]
\centering
\caption{Cross-dataset reliability summary. ``--'' = not natively produced by the dataset's evaluation protocol; ``$\dagger$'' indicates a single split (no fold variance). All numbers come from the corresponding dataset-specific tables earlier in the paper / appendix.}
\label{tab:reliability-summary}
\scriptsize
\setlength{\tabcolsep}{4pt}
\renewcommand{\arraystretch}{1.15}
\resizebox{\textwidth}{!}{%
\begin{tabular}{l l l l l}
\toprule
\textbf{Dataset} & \textbf{Source of variability} & \textbf{Variability magnitude} & \textbf{Calibration / coverage} & \textbf{Reliability mechanism this paper exposes} \\
\midrule
PPMI    & 6-model architectural ablation & MAE range $4.579$--$4.918$ (Table~\ref{tab:ppmi-ablation}) & --$\dagger$ (regression) & Node-attention temporal interval (interpretation only) \\
mPower  & 5 random seeds (ensemble)      & per-seed AUC $0.945$--$0.952$, $\sigma=0.003$ (Table~\ref{tab:mpower-seed-ensemble}) & val-selected threshold = $0.75$, frozen on test & Ensemble variance + best-F1 acceptance threshold \\
PADS    & 5-fold subject CV + 50-subj blind cohort & visible $\sigma_{\mathrm{AUC}}=0.029$; blind $\Delta\mathrm{ACC}=-0.064$ (NS, 95\% CIs overlap, Table~\ref{tab:pads-hidden}) & ECE $0.100\to 0.161$, Brier $0.473\to 0.500$, ICC $0.902\to 0.600$, conformal $0.804\pm 0.053$ (target $0.80$) & Quality + uncertainty + OOD; 4-action router (Fig.~\ref{fig:pads-routing}) \\
Daphnet & 6 LOSO subject folds           & F1 $\sigma=0.080$, AUC $\sigma=0.038$, AUPRC $\sigma=0.057$; range F1 $0.674$--$0.922$ (Table~\ref{tab:daphnet-loso}) & --$\dagger$ (per-fold accuracy / F1 only) & LOSO subject-level variance \\
UCI     & 5 temporal-anchor folds        & motor MAE $8.38\to 3.24$ over $n_{\mathrm{anchor}}\in\{0,5,10,20\}$ (Table~\ref{tab:uci-personalized-app}) & split-conformal coverage motor $0.81$--$0.99$, total $0.55$--$0.99$ at target $0.80$ (per-fold) & Selective-MAE 4-action router; anchor-sufficiency signal \\
\bottomrule
\end{tabular}%
}
\end{table}

The summary clarifies the actual reach of the reliability stack: per-branch calibration / conformal coverage / 4-action routing are demonstrated in full on PADS and (in regression form) on UCI; Daphnet and mPower expose subject-level / seed-level variance evidence; PPMI provides regression-architecture ablation evidence. We do not claim that calibration metrics such as ECE were measured uniformly across all five datasets; doing so on the full set is left as an explicit future-work item.

\subsection{Reliability calibration and uncertainty estimation}

The reliability estimation, conformal prediction, per-branch calibration, and selective rejection mechanism introduced by SGC-RML achieve accurate uncertainty quantification and prediction calibration. Experimental results show that the model's conformal coverage reaches $0.804\pm0.053$, highly consistent with the target coverage of $0.80$, proving that the model's uncertainty estimation has excellent calibrability and can accurately determine the credibility of the prediction results. Meanwhile, the personalized adaptation experiment on the UCI dataset (Table~\ref{tab:uci-personalized-app}) verified the framework's adaptive reliability control capability: when evidence is insufficient (no anchored data), the model outputs low-confidence results, which can trigger active rejection and retesting; when evidence is sufficient, it achieves high-accuracy prediction, supporting the rigour of retrospective assessment from a mechanism perspective.

\paragraph{Per-metric definitions and the PADS hidden-blind degradation pattern.}
ECE is the expected calibration error calculated based on a 15-bin partition after temperature calibration~\cite{guo2017,naeini2015obtaining}; Brier is the multi-classification form $\frac{1}{N}\sum_{i,c}(\hat{p}_{i,c}-y_{i,c})^2$ (smaller is better); ICC uses ICC(2,1) absolute consistency~\cite{shrout1979}. Under PADS hidden-blind testing (Table~\ref{tab:pads-hidden}), ECE increased from $0.100$ to $0.161$, Brier from $0.473$ to $0.500$, and ICC dropped from $0.902$ to $0.600$. This shows that out-of-distribution or absent subjects do indeed lead to calibration degradation and decreased consistency. The phenomenon is not a failure but a necessary observation: if only AUC were reported, the model would appear nearly risk-free; adding ECE / Brier / ICC reveals that probabilistic calibration and consistency carry deployment risks even when ranking remains strong. The conformal coverage results corroborate this: the empirical coverage of the hidden blind test was $0.804 \pm 0.053$, close to the $0.80$ target; the visible 5-fold coverage was $0.824$. Even with degraded calibration metrics on hidden subjects, conformal prediction maintained near-target coverage, directly supporting SGC-RML's claim of ``knowing when to be reliable.''

\paragraph{UCI anchoring and triggered test-retesting.}
From a retrospective clinical assessment perspective, the UCI anchor finding (Table~\ref{tab:uci-personalized-app}) shows that the system does not require large amounts of individual data to achieve benefits; a small number of anchors can substantially reduce errors. This provides a reasonable basis for ``triggered test-retesting'': when the model finds that the current input deviates too much from the individual anchor, the evidence is insufficient, or the uncertainty is too high, test-retesting is not an add-on feature but rather part of a reliable longitudinal assessment process.

\subsection{Clinical interpretability validation}

SGC-RML's symptom-node-level attention mechanism produces an interpretation that is consistent with the clinical assessment system. Experimental results show that the model's cumulative attention ratio on the three core motor symptoms (tremor, bradykinesia, and gait abnormalities) of UPDRS-III reaches $95\%$, in line with the core dimensions of clinical PD assessment. The full per-node attention distribution is reported in Table~\ref{tab:ppmi-attention}; the attention pattern is automatically learned during training rather than hard-coded, and low-weight non-motor nodes (mood, sleep/autonomic) match the expected weak direct contribution to UPDRS-III motor scores. Note, however, that attention weights provide explanatory hints rather than causal evidence~\cite{jain2019,wiegreffe2019}, and spurious correlations cannot be strictly ruled out. This result indicates that the model's prediction is not a black-box output but rather inference grounded in clinically recognized core symptom dimensions, providing clinicians with visualized and interpretable symptom-level evidence.

\begin{table}[H]
\centering
\caption{PPMI per-node mean attention (top-3 \textbf{bolded}). Attention weights are interpretation hints, not causal evidence~\cite{jain2019,wiegreffe2019}.}
\label{tab:ppmi-attention}
\small
\setlength{\tabcolsep}{4pt}
\resizebox{\textwidth}{!}{%
\begin{tabular}{c l r r p{6.5cm}}
\toprule
\textbf{Rank} & \textbf{Symptom node} & \textbf{Attention} & \textbf{Cum.} & \textbf{Clinical interpretation} \\
\midrule
  1 & \textbf{bradykinesia} & \textbf{0.398} & \textbf{0.398} & Cardinal motor symptom (UPDRS-III dominant) \\
  2 & \textbf{tremor} & \textbf{0.368} & \textbf{0.766} & Cardinal motor symptom \\
  3 & \textbf{motor\_fluctuation} & \textbf{0.187} & \textbf{0.953} & Disease progression marker \\
  4 & cognition & 0.027 & 0.980 & Non-motor (less weighted in UPDRS-III) \\
  5 & sleep\_autonomic & 0.010 & 0.990 & Non-motor \\
  6 & reliability\_state & 0.007 & 0.997 & Auxiliary reliability signal (not a clinical symptom) \\
  7 & axial\_gait & 0.003 & 1.000 & Underrepresented in tabular UPDRS-III \\
  8 & mood & 0.001 & 1.000 & Non-motor (lowest weight) \\
\bottomrule
\end{tabular}%
}
\end{table}

\medskip
\noindent
Overall, the existing results support the core argument of this paper: SGC-RML advances digital assessment of PD from simply optimizing average prediction accuracy to framework-level modeling that simultaneously considers cross-modal evidence integration, individualized adaptation, reliability estimation, and clinical interpretability. For real-world digital neural systems, this capability is more important than small performance improvements on a single dataset and is closer to the actual needs of clinically-assisted assessment systems.

\section{Reproducibility card}
\label{app:repro-card}

Table~\ref{tab:repro-card} consolidates per-dataset training and evaluation specifications. Anonymous code release is provided as supplementary material; runtime reflects single-GPU wall-clock on an NVIDIA A40 (48\,GB) on an internal shared cluster.

\begin{table}[H]
\centering
\caption{Per-dataset reproducibility card. ``Threshold'' is the per-sample decision threshold used by the routing layer (acceptance threshold for classification, conformal residual quantile $\hat{q}$ for regression).}
\label{tab:repro-card}
\scriptsize
\setlength{\tabcolsep}{3pt}
\renewcommand{\arraystretch}{1.20}
\resizebox{\textwidth}{!}{%
\begin{tabular}{l l l l l l l l l}
\toprule
\textbf{Dataset} & \textbf{Encoder} & \textbf{Split} & \textbf{Optim.\,/\,LR} & \textbf{Batch} & \textbf{Epochs / ES} & \textbf{Seeds} & \textbf{Threshold} & \textbf{Runtime} \\
\midrule
PPMI    & Node Attention TempT.   & patient-level 80/20  & AdamW $8\!\cdot\!10^{-4}$ & 64  & 100 / patience 15  & 1 (deterministic init) & N/A (regression)              & $\sim$1\,d \\
mPower  & FT-Transformer ensemble & subject-indep.\ 5fCV & Adam   $10^{-3}$           & 256 & 100 / patience 10  & 5 (\{42, 2024, 3407, 7, 123\}) & val-selected best-F1 = $0.75$ & $\sim$1.5\,d \\
PADS    & PatchTST + Quality head & subject 5fCV + 50-blind & AdamW $5\!\cdot\!10^{-4}$ & 32 & 80 / patience 12 & 3 (V7-s\{0,1,2\}) & val-quantile thresholds (\S\ref{sec:reliability-routing}) & $\sim$2\,d \\
Daphnet & TimeSeries Transformer  & 6 LOSO + S09 fixed val & AdamW $10^{-3}$           & 64  & 50 / patience 8    & 1 per fold              & best-F1 per fold              & $\sim$1\,d \\
UCI     & FT-Transformer + personalized & temporal-anchor 5fCV & Adam $10^{-3}$           & 128 & 60 / patience 10   & 1                       & conformal $\hat{q}$, $\alpha=0.2$ & $\sim$0.5\,d \\
\bottomrule
\end{tabular}%
}
\end{table}

\paragraph{Anonymized supplementary code.}
A self-contained code release accompanies this submission, structured as
\verb|core/|, \verb|pads/|, \verb|uci/|, \verb|mpower/|, \verb|ppmi/|, \verb|daphnet/|.
All paths and identifiers in the source have been anonymized. The
\verb|core/reliability/| package implements the conformal coverage and 4-action
router used by every dataset; per-dataset training scripts under each
\verb|<dataset>/scripts/| reproduce the corresponding tables in this paper from the
public datasets cited in Section~\ref{sec:method}.

\input{checklist.tex}

\end{document}

%% file: SGC_RML_section1_intro.tex

\section{Introduction}
\label{sec:intro}

Parkinson's disease (PD) is the second most prevalent and fastest-growing neurological disorder worldwide, with an estimated 25 million patients by 2050~\cite{dorsey2018,su2025,bloem2021parkinson}. Current clinical assessments primarily rely on scales such as the MDS-UPDRS and periodic outpatient examinations; however, these assessments are subjective, infrequent, and context-dependent, making it challenging to capture patients' symptom fluctuations, drug responses, and longitudinal progression in real-life settings~\cite{goetz2008,latourelle2017large}. With the development of smartphones, wearable devices, voice acquisition, gait sensors, and remote follow-up systems, PD assessment is shifting from low-frequency in-hospital assessments to high-frequency digital measurements in natural environments~\cite{espay2016technology,patel2012review,acosta2022multimodal}. However, this shift has not simply created a larger-scale supervised learning problem but rather exposed more complex machine learning challenges: real-world digital neural data is often acquired across devices, centers, tasks, and time periods and is accompanied by modality loss, signal noise, irregular longitudinal sampling, incomplete labels, and subject distribution bias. Therefore, the key to digital assessment of PD is not just improving average predictive performance but learning transferable symptom representations under heterogeneous and incomplete evidence, judging when the model is reliable and when it is not.

Existing machine learning methods often target average performance metrics such as AUC, F1, MAE, or R$^2$, optimizing models on a single dataset, in a single modality, or under a fixed protocol. While these studies have promoted the application of smart devices and machine learning in PD assessment, there is still a significant gap between them and retrospective reliability-aware assessment for real-world clinical workflows. First, multimodal models often assume that the input is relatively complete, while real-world acquisition commonly suffers from problems such as sensor detachment, incomplete tasks, unstable speech quality, and uneven follow-up intervals~\cite{baltrusaitis2019,wu2024}. Second, existing models are usually forced to give deterministic outputs, lacking the ability to distinguish between decision states such as ``predictable,'' ``low-confidence aid,'' ``retest required,'' or ``should be rejected.'' Finally, interpretation methods mostly remain at the level of low-level feature contributions and cannot yet stably map to the clinical symptom structures such as tremor, bradykinesia, gait abnormalities, frozen gait, cognition, and emotion. Therefore, reliability-oriented digital assessment of PD needs to shift from ``always predicting'' to ``conditionally reliable predicting,'' while possessing the ability to calibrate uncertainty, guarantee coverage, selectively reject, and provide symptom-level interpretation~\cite{baltrusaitis2019,wu2024,guo2017,vovk2005,angelopoulos2023,geifman2019b}.

Therefore, we propose the Symptom-Graph-Conditioned Reliable Multimodal Learning (SGC-RML) framework for reliable and interpretable longitudinal assessment of PD in real-world digital neural systems. SGC-RML avoids simple splicing or posterior fusion in the original modal space, instead mapping heterogeneous observations such as speech, gait, wearable sensors, smartphone tasks, and clinical follow-ups to a shared clinical symptom atlas space. Specifically, the framework constructs symptom nodes based on clinical dimensions such as tremor, bradykinesia, gait abnormalities, frozen gait, cognition, emotion, and autonomic symptoms, and models inter-symptom dependencies through symptom graph conditional representation. Building upon this, SGC-RML introduces per-branch calibration, conformal prediction, and selective rejection mechanisms, enabling the model to not only output symptom or severity estimates but also assess the reliability of predictions and reject high-risk outputs or suggest re-collection when evidence is insufficient. Simultaneously, the framework generates evidence summaries corresponding to clinical structures through symptom node-level attention, interpreting multimodal signals as readable symptom contributions rather than isolated low-level feature importance.

This paper makes three main contributions:
\begin{enumerate}
\item \textbf{Routing in clinical symptom space, not raw feature space.} We project heterogeneous PD signals (speech, IMU, mobile tasks, clinical records) onto a shared 8-dim symptom-node space (7 clinical nodes plus a \texttt{reliability\_state} auxiliary) so that the same per-branch calibration, missingness mask, and decision routing apply uniformly across all five datasets.
\item \textbf{Four routing actions tied to distinct uncertainty sources.} Unlike single-threshold reject options in prior digital-biomarker work, our deterministic 4-action policy ties each action to a \emph{different} signal---acquisition incompleteness drives \textsc{reacquire}, predictive uncertainty drives \textsc{abstain}, OOD score drives \textsc{refer}, calibrated interval width gates \textsc{predict}---and is fixed by validation-frozen quantiles, reused across all datasets without retuning.
\item \textbf{Cross-dataset reliability evidence under realistic protocols.} We validate the framework on five PD datasets (mPower, PADS, PPMI, Daphnet FoG, UCI Telemonitoring) spanning classification, regression, event detection, and longitudinal severity prediction, including a hidden 50-subject blind PADS cohort and leak-free temporal anchoring on UCI. The reliability stack holds conformal coverage near the 0.80 target even when point-prediction calibration degrades on hidden subjects---evidence that the stack, not the point predictor, is what generalises.
\end{enumerate}

%% file: SGC_RML_section2_related.tex

\section{Related work}
\label{sec:related}

In recent years, digital assessment of PD has shifted from single digital biomarker modeling to real-world, multi-source, and longitudinal monitoring. Early studies primarily relied on speech, gait, or wearable acceleration signals for diagnosis, severity estimation, and frozen gait detection~\cite{little2009suitability,sakar2013collection,patel2012review,camps2018deep,sigcha2020deep}. Subsequently, mPower validated the feasibility of remotely acquiring large-scale PD motor behavior data via smartphones, UCI Parkinson's Telemonitoring demonstrated the value of estimating motor/total UPDRS using home speech signals, Daphnet FoG advanced wearable frozen gait detection research, PPMI provided a multi-center, longitudinal paradigm combining clinical assessment and biomarkers for disease progression research, and PADS introduced dual-wrist smartwatches, multi-task motor evoked data, and symptom-labeled data, laying the foundation for cross-device, cross-task digital motor phenotypic modeling~\cite{bot2016,marek2011,tsanas2010,bachlin2010,varghese2024}. These resources have driven the application of machine learning and deep learning in PD diagnosis, symptom assessment, remote monitoring, and progression modeling~\cite{lonini2018,moreau2023,latourelle2017large,espay2016technology}. However, they also indicate that real-world digital PD data is not clean, complete, or single-task data, but rather a complex clinical data stream composed of multimodal missing data, noise, device differences, centrality differences, and irregular longitudinal observations~\cite{acosta2022multimodal,finlayson2021clinician}.

For reliability modeling, probabilistic calibration, Bayesian uncertainty, deep ensembles, conformal prediction, and selective classification mitigate neural-network overconfidence, build prediction intervals, or reject high-risk samples~\cite{angelopoulos2023,kendall2017,geifman2017,gal2016dropout,lakshminarayanan2017simple,el-yaniv2010foundations,hendrycks2017baseline,ovadia2019can,tibshirani2019conformal}. However, these are typically task-agnostic post-processing modules disconnected from PD symptom structure, modality loss, and longitudinal change mechanisms, and rarely translate into clinically actionable decisions (accept, downgrade, retest, reject). Reliability in PD should therefore be organised at the symptom, modality, and action level, not as a single global confidence score.

Interpretive learning offers complementary tools: LIME, SHAP, and attention mechanisms expose feature, time-segment, or modality contributions~\cite{ribeiro2016,lundberg2017}, but general attribution is not clinical evidence and attention weights are not explanations~\cite{jain2019}. Useful PD interpretation should map modal evidence to specific symptom nodes (tremor, bradykinesia, gait, speech, cognition, autonomic) and judge whether this evidence is sufficient under current reliability conditions.
Based on the aforementioned gaps, SGC-RML is not another PD predictor that only pursues average performance on a single dataset, but rather a reliable multimodal learning framework for real-world digital neural systems. Compared to existing methods, SGC-RML advances the question from ``can it predict?'' to ``when is the model reliable, why is it reliable, and how should we act when it is unreliable?''

%% file: SGC_RML_section3_method.tex

\section{Method}
\label{sec:method}

\subsection{Benchmark and unified formulation}
\label{sec:benchmark}

We evaluate SGC-RML on five public PD datasets spanning wearable motor sensing, remote smartphone assessment, longitudinal clinical records, freezing-of-gait detection, and voice-based telemonitoring. Rather than enforcing a shared raw-input format, dataset-specific encoders project modality-appropriate evidence into a common 8-dim symptom-node representation; the shared module then performs symptom-level alignment, calibration, and reliability routing under a unified prediction-and-decision protocol (Figure~\ref{fig:framework}; per-dataset feature-to-node mapping in Appendix~\ref{app:node-mapping}, Table~\ref{tab:feature_node_mapping}).

\begin{table}[!ht]
\centering
\caption{Summary of the datasets.}
\label{tab:dataset_statistics}
\scriptsize
\setlength{\tabcolsep}{4pt}
\renewcommand{\arraystretch}{1.08}
\begin{tabular*}{\textwidth}{@{\extracolsep{\fill}} l c c c c c c c @{}}
\toprule
\textbf{Dataset} & \textbf{\#Subjects} & \textbf{\#Targets} & \textbf{Site} & \textbf{Device} & \textbf{Temporal} & \textbf{Data} & \textbf{Issue} \\
\midrule
PADS & 469 & 3 & 1 clinic & Watch+phone & 1 visit & IMU & QC missing \\
mPower & 8{,}320 & 2 & Remote & iPhone & Repeated & Phone tasks/Survey/Demo. & Missing \\
PPMI & 5{,}500 & 2 & 50 sites & Clinical & Longitudinal & UPDRS/Scales/Gait-med. & Missing \\
Daphnet FoG & 10 & 2 & Lab & 3 acc. & Task-based & Accelerometer & Imbalanced \\
UCI Telemonitoring & 42 & 2 & Home & Voice recorder & 6 months & Voice & Sparse cohort \\
\bottomrule
\end{tabular*}
\vspace{2pt}
\begin{minipage}{\textwidth}
\scriptsize
\textit{Notes.} ``Data'' denotes the modality or feature group used in our pipeline. Phone tasks include tapping, walking, voice, and memory tasks; ``Demo.'' denotes demographic or task-availability variables. Scales include MoCA, SCOPA-AUT, GDS, RBD, and UPSIT; Gait-med. includes FoG, falls, and medication-related variables. ``acc.'' = accelerometers. ``Issue'' denotes the dominant data-quality challenge.
\end{minipage}
\end{table}
\FloatBarrier
\begin{wrapfigure}[12]{l}{0.50\textwidth}
\vspace{-1.0\baselineskip}
\centering
\includegraphics[width=0.50\textwidth]{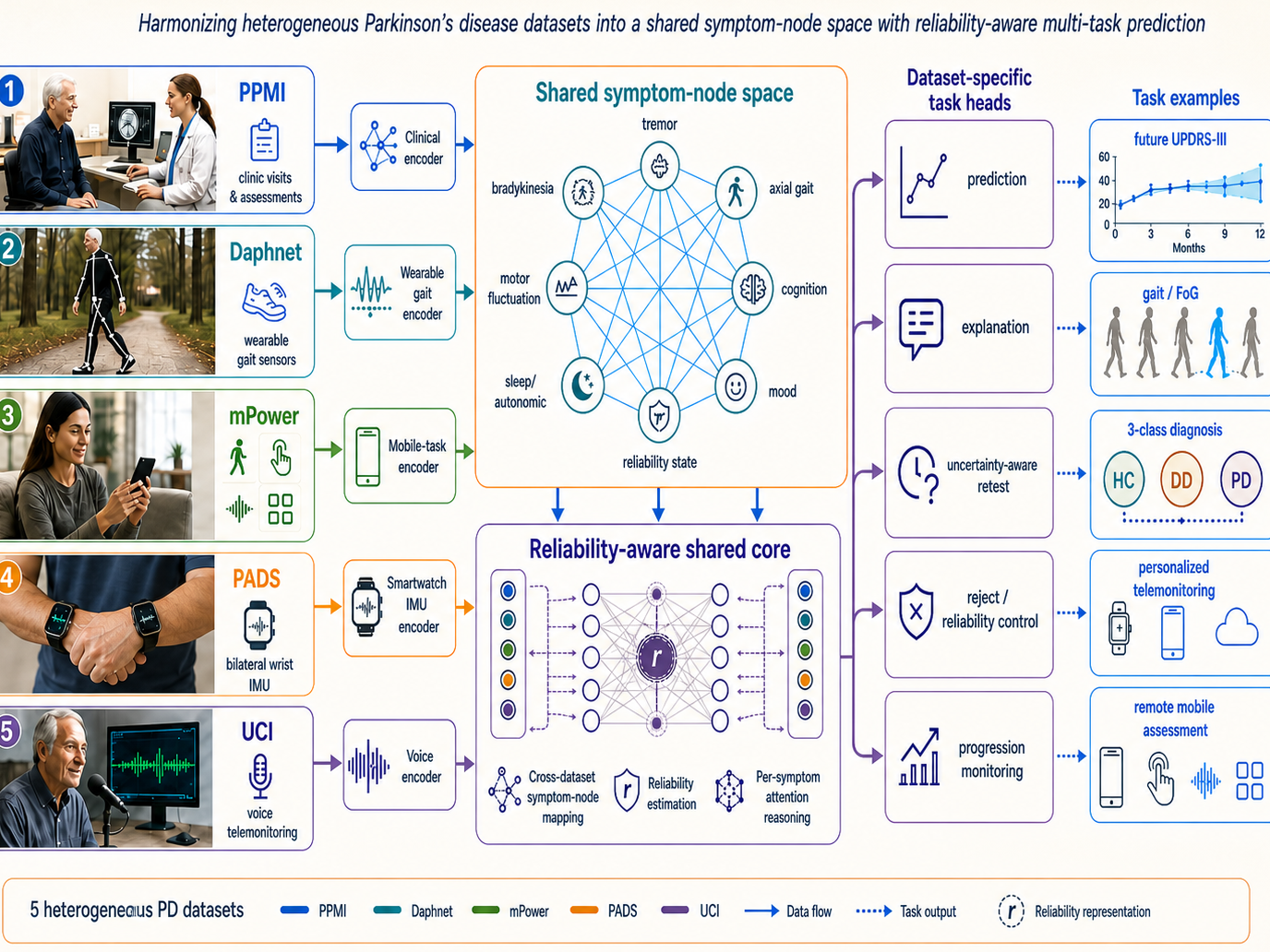}
\caption{\textbf{Overview of SGC-RML.} Heterogeneous PD datasets are routed via a shared 8-dim symptom-node space and a reliability-aware 4-action core.}
\label{fig:framework}
\vspace{-0.6\baselineskip}
\end{wrapfigure}
\paragraph{Unified encoder--head--symptom interface.}
For a sample $x$ from dataset $d$, SGC-RML applies a dataset-specific encoder $f_d$, a dataset-specific prediction head $g_d$, and a symptom projection module $\phi_d$:
\begin{equation}
    h_d = f_d(x), \qquad
    \hat{y} = g_d(h_d), \qquad
    s_d = \phi_d(h_d).
    \label{eq:encoder-head-translation}
\end{equation}
The encoder and prediction head are modality-specific, while symptom projection, calibration, and reliability routing follow a shared protocol across datasets.

\clearpage
\paragraph{Shared 8-dimensional symptom-node space.}
The shared clinical schema contains seven symptom nodes plus one auxiliary \textsc{reliability\_state} node:
\begin{equation}
\mathbf{s}_i =
\big[
s_{\mathrm{tremor}},
s_{\mathrm{brady}},
s_{\mathrm{gait}},
s_{\mathrm{fluct}},
s_{\mathrm{cog}},
s_{\mathrm{sleep/aut}},
s_{\mathrm{mood}},
s_{\mathrm{rel}}
\big]
\in [0,1]^8 .
\label{eq:symptom-vector}
\end{equation}
The seven clinical nodes represent tremor, bradykinesia, axial/gait impairment, motor fluctuation, cognition, sleep/autonomic symptoms, and mood. Heterogeneous clinical scales are normalized into the unit interval, and indicators where higher values reflect better function are flipped into severity scores:
\begin{equation}
x_{\mathrm{norm}} =
\frac{x - x_{\min}}{x_{\max} - x_{\min} + \epsilon},
\qquad
s_{\mathrm{severity}} = 1 - x_{\mathrm{norm}} .
\label{eq:minmax-severity}
\end{equation}

\paragraph{Missingness-aware reliability state.}
Different datasets observe different subsets of symptom nodes. We therefore retain both the symptom value and an explicit missingness indicator
$m_i^{(j)} = \mathbb{I}\!\left(s_i^{(j)} \text{ is missing}\right)$
for each node. The auxiliary reliability state is not a clinical symptom; it summarizes whether the current sample is sufficiently complete, trustworthy, and certain for downstream use:
\begin{equation}
r_i = f\!\left(\text{completeness}_i,\; \text{quality}_i,\; \text{uncertainty}_i\right).
\label{eq:reliability}
\end{equation}

\subsection{Dataset-specific evidence extraction}
\label{sec:evidence}

Each dataset uses a modality-appropriate encoder: a patch-based Transformer for PADS IMU tasks, an FT-Transformer for UCI voice-derived tabular features, a temporal Transformer for PPMI longitudinal visits, an FT-Transformer ensemble for mPower digital tasks, and a time-series Transformer for Daphnet FoG windows. All encoders share the self-attention backbone of~\cite{vaswani2017attention}; the time-series variants follow the design space surveyed in~\cite{wen2023transformers}. Dataset-specific losses, windowing details, and feature-to-node mappings are provided in Appendix~\ref{app:dataset-details}--\ref{app:node-mapping}.

\paragraph{PADS.}
PADS contains two-wrist six-channel IMU recordings from 469 subjects performing 11 motor tasks~\cite{varghese2024}. We use a patch-based time-series Transformer~\cite{nie2023patchtst} with task and wrist embeddings for PD/DD/HC classification, followed by subject-level aggregation across tasks and wrists.

\paragraph{UCI Telemonitoring.}
UCI Telemonitoring provides repeated voice-derived features with demographic and temporal covariates for motor and total UPDRS regression~\cite{tsanas2010}, building on earlier dysphonia-based PD telemonitoring evidence~\cite{little2009suitability,sakar2013collection}. An FT-Transformer~\cite{gorishniy2021revisiting} encodes the tabular voice biomarkers, and quantile outputs provide uncertainty-aware severity estimates. A personalised variant predicts residual severity relative to early-visit subject anchors.

\paragraph{PPMI.}
PPMI is used as the main longitudinal clinical branch~\cite{marek2011}. Visit-level records include UPDRS scores, non-motor scales, olfaction, gait/freezing records, falls, and medication-related variables. For patient $p$, the visit history is represented as
$\mathbf{X}_p = [x_{p,1}, \ldots, x_{p,T}]$, and the primary target is future motor severity:
\begin{equation}
y_{p,t} =
\mathrm{UPDRS\text{-}III}_p(t + 12\ \mathrm{months}),
\qquad
\hat{y}_p =
g_\theta\!\left(\mathrm{Transformer}_\theta(\mathbf{X}_p)\right).
\label{eq:ppmi-prediction}
\end{equation}
Progression-related auxiliary labels are derived from 12-month UPDRS-III change and long-term progression slope when follow-up is sufficient. Optional node-attention analysis is reported in Appendix~\ref{app:dataset-details}.

\paragraph{mPower.}
mPower is treated as a mobile digital-biomarker branch for PD/HC classification~\cite{bot2016}. Its input combines tapping, walking, voice, memory, demographic, and task-availability variables. An FT-Transformer ensemble encodes task-specific digital features. The classification objective and feature-to-node mapping are described in Appendix~\ref{app:dataset-details}--\ref{app:node-mapping}.

\paragraph{Daphnet FoG.}
Daphnet FoG is modeled as a wearable gait/freezing branch~\cite{roggen2013daphnet}. We segment accelerometer streams into overlapping windows and train a time-series Transformer to predict the probability of freezing of gait for each window, building on prior deep-learning FoG detectors~\cite{camps2018deep,sigcha2020deep}. Windowing details are given in Appendix~\ref{app:dataset-details}.

\subsection{Calibration and reliability routing}
\label{sec:reliability-routing}

\paragraph{Calibration.}
Confidence values from different branches are not directly comparable, so each branch is calibrated before reliability estimation. Classification branches use temperature scaling, while regression branches use quantile or split-conformal calibration. For regression, the calibration residual is defined as
\begin{equation}
E_i =
\max\!\big(
\hat{q}_{\mathrm{low},i} - y_i,\;
y_i - \hat{q}_{\mathrm{high},i}
\big),
\label{eq:conformal-residual}
\end{equation}
and is used to widen prediction intervals so that empirical coverage matches the target level.

\paragraph{Four-action decision routing.}
The reliability module combines acquisition quality, predictive uncertainty, missingness, and out-of-distribution evidence~\cite{hendrycks2017baseline}, and outputs one of four actions:
\begin{equation}
\mathcal{A}
=
\{
\textsc{Predict},
\textsc{Abstain},
\textsc{Reacquire},
\textsc{Refer}
\}.
\label{eq:action-set}
\end{equation}
\textsc{Predict} accepts reliable and low-uncertainty outputs; \textsc{Abstain} withholds uncertain predictions; \textsc{Reacquire} flags poor-quality or incomplete inputs; and \textsc{Refer} is used for clinically difficult or distributionally atypical cases.

\paragraph{Deterministic post-hoc policy.}
To make routing reproducible, we implement the decision module as a deterministic post-hoc policy shared across all datasets. For each sample $i$, the router takes the symptom-node vector $s_i$, missingness mask $m_i$, calibrated prediction $\hat{y}_i$, acquisition-quality score $q_i$, predictive uncertainty $u_i$, and out-of-distribution score $o_i$. We first compute a dataset-aware completeness score:
\begin{equation}
c_i =
1 -
\frac{1}{|O_d|}
\sum_{j\in O_d} m_i^{(j)},
\label{eq:completeness}
\end{equation}
where $O_d$ denotes the symptom nodes observable in dataset $d$. This prevents a modality from being penalized for nodes it cannot capture, such as cognition or mood in Daphnet FoG.

The final action is selected by a fixed priority rule:
\begin{equation}
A_i =
\begin{cases}
\textsc{Reacquire}, & q_i < \tau_q \ \text{or}\ c_i < \tau_c, \\
\textsc{Refer}, & o_i > \tau_{\mathrm{ood}}, \\
\textsc{Abstain}, & u_i > \tau_u \ \text{or}\ \mathrm{width}(C_i) > \tau_w, \\
\textsc{Predict}, & \text{otherwise}.
\end{cases}
\label{eq:routing-rule}
\end{equation}
For classification, $C_i$ denotes the conformal prediction set~\cite{romano2019conformalized}, and $\mathrm{width}(C_i)$ can be replaced by set size or predictive entropy. For regression, $C_i$ denotes the calibrated prediction interval. All thresholds are selected on the validation split as fixed quantiles of the per-sample signals and frozen before held-out test evaluation. Concretely, on PADS, $\tau_q$ is set to the $20{\rm th}$ percentile of the validation quality score ($\tau_q = 0.9816$), $\tau_u$ for the $\textsc{abstain}$ rule is the $70{\rm th}$ percentile of validation uncertainty ($\tau_u^{\mathrm{abstain}} = 0.2182$), and the $\textsc{refer}$ rule uses the $85{\rm th}$ percentile of uncertainty ($\tau_u^{\mathrm{refer}} = 0.2427$). The OOD threshold $\tau_{\mathrm{ood}}$ is the $85{\rm th}$ percentile of the validation OOD score ($\tau_{\mathrm{ood}} = 0.5706$). For regression branches, $\tau_w$ is set so that the validation prediction-set width quantile equals the desired coverage budget; on UCI Telemonitoring this resolves to $\tau_w \approx 0.75$ (motor) and $\tau_w \approx 0.80$ (total). The completeness threshold $\tau_c$ is dataset-specific: $\tau_c = |O_d|/|O_d|$ for single-dataset evaluation (no within-dataset node missingness) and $\tau_c = 0.5$ across the cross-dataset routing protocol. Thresholds are deterministic given the validation split and are not retuned per fold.

\paragraph{What is technically new.}
SGC-RML re-uses standard calibration and conformal-prediction primitives, but the contribution is the way they are composed into an auditable retrospective decision interface for heterogeneous PD assessment. Concretely, SGC-RML differs from generic post-hoc calibration or selective classification in three specific ways: \emph{(i)} routing decisions are taken in the shared 8-dim symptom-node space rather than in raw feature space, allowing the same per-branch calibration and per-node missingness mask to apply uniformly across speech, IMU, mobile-task, and clinical-record modalities; \emph{(ii)} the four actions are tied to \emph{distinct} sources of uncertainty---acquisition incompleteness ($q_i, c_i$) drives \textsc{Reacquire}, predictive uncertainty ($u_i$) and calibrated interval width drive \textsc{Abstain}, distributional atypicality ($o_i$) drives \textsc{Refer}---rather than collapsing all reliability concerns into one confidence score; and \emph{(iii)} the policy is a deterministic, validation-frozen quantile rule (Eq.~\ref{eq:routing-rule}) that admits a per-sample audit trail of which signal triggered the routing outcome. We do not claim a new uncertainty estimator; we claim a symptom-structured, action-level, auditable decision interface for retrospective reliability-aware assessment of digital PD measurements.

\FloatBarrier

%% file: SGC_RML_section4_full.tex

\setlength{\intextsep}{6pt plus 2pt minus 2pt}
\setlength{\textfloatsep}{6pt plus 2pt minus 2pt}
\setlength{\abovecaptionskip}{4pt}
\setlength{\belowcaptionskip}{2pt}

\section{Experiments}
\label{sec:experiments}

\subsection{Experimental setup}
\label{sec:exp-setup}
This section systematically evaluates the SGC-RML on five real-world heterogeneous PD datasets, covering five clinical tasks: longitudinal regression of UPDRS-III 12-month future results, binary classification of PD/HC, triclass classification of HC/PD/atypical Parkinson's syndrome, FoG event detection, and personalized UPDRS prediction. This cross-dataset, cross-task experimental design was used to examine the framework's comprehensive capabilities in predictive performance, generalization robustness, clinical interpretability, uncertainty calibration, and personalized adaptation. All experiments employed rigorous evaluation protocols closely aligned with retrospective clinical assessment workflows: PPMI used patient-level longitudinal splitting (each subject appears in exactly one of train / val / test, preventing within-subject temporal leakage); mPower used subject-independent 5-fold cross-validation; Daphnet used LOSO; PADS used a 50-subject fully independent blinded test set held out from training, validation, and model selection; and UCI used leak-free temporal anchoring. This design minimized the risk of data leakage and supports retrospective reproducibility of the results.

\subsection{Overall performance across five real-world PD datasets}
\label{sec:exp-overall}
Fig.~\ref{fig:main-cards} and Table~\ref{tab:main} show the overall performance of SGC-RML on five heterogeneous real-world Parkinson's disease datasets. Unlike evaluations of a single task or a single modality, this experiment covered multiple task types, including UPDRS-III regression over the next 12 months, PD/HC binary classification, HC/PD/DD tri-class classification, frozen gait event detection, and personalized UPDRS longitudinal prediction. Therefore, this result primarily validates not the local optima of a specific model structure, but rather the applicability of SGC-RML as a unified reliability framework across multiple tasks, modalities, and data distributions.


\begin{table}[!ht]
\centering
\caption{\textbf{Main results.} Best per-dataset method and primary metric.
\textbf{Bold} = best on that dataset; protocols summarised in
\S\ref{sec:exp-setup}.
\textsuperscript{a}~PADS reports two operating points: visible 5-fold CV (n=419, AUC 0.805) and held-out 50-subject blind cohort (AUC 0.825); see Table~\ref{tab:pads-hidden}.
\textsuperscript{b}~Daphnet contains 10 subjects (S01--S10); S09 is held as the fixed validation set; six LOSO folds (S01--S06) completed within our compute budget. Reported F1/AUC are mean over these six folds.}
\label{tab:main}
\small
\setlength{\tabcolsep}{4pt}
\resizebox{\textwidth}{!}{%
\begin{tabular}{l r l l l l}
\toprule
\textbf{Dataset} & $n_{\text{test}}$ & \textbf{Task type} & \textbf{Best method} & \textbf{Primary} & \textbf{Secondary} \\
\midrule
  \textbf{PPMI} & 3,376 & future 12m UPDRS-III regression & Node Attention Temporal Transformer & \textbf{MAE = 4.579} & \textbf{R2 = 0.772} \\
  \textbf{mPower} & 970 & binary PD/HC classification & FT-Transformer Ensemble (5 seeds) & \textbf{AUC = 0.953} & \textbf{F1 = 0.749} \\
  \textbf{PADS} & 419+50 & 3-class (HC/PD/DD); 50-subj hidden blind & V7 Ensemble (PatchTST + Quality) & \textbf{AUC = 0.805 / 0.825}\textsuperscript{a} & \textbf{F1 = 0.588} \\
  \textbf{Daphnet} & 6 & FoG event detection (6-fold LOSO)\textsuperscript{b} & TimeSeries Transformer & \textbf{F1 = 0.803} & \textbf{AUC = 0.872} \\
  \textbf{UCI} & 5,875 & personalized UPDRS regression (n=20 anchor) & FT-Transformer + Personalized & \textbf{MAE = 3.240} & \textbf{CCC = 0.756} \\
\bottomrule
\end{tabular}%
}
\end{table}

Across the five datasets, SGC-RML achieves competitive performance under task-specific protocols while sharing the same symptom-node and reliability interface (Table~\ref{tab:main}). PADS in particular shows AUC stability between visible 5-fold CV (0.805) and the 50-subject hidden blind cohort (0.825); per-dataset details are reported in the dedicated subsections \S\ref{sec:exp-pads-hidden} and \S\ref{sec:exp-uci-personalized}. We do not use these heterogeneous tasks for direct cross-dataset numerical ranking; the goal is to evaluate whether the same reliability stack can be applied across modalities and objectives.

The contribution is therefore a shared reliability stack across dataset-specific encoders, not a shared encoder: symptom maps, calibration, conformal coverage, and 4-action routing are reused across modalities while the optimal backbone is retained for each task.

\subsection{The shared symptom-node space supports cross-dataset alignment}
\label{sec:exp-symptom-graph}
\enlargethispage{2\baselineskip}

\begin{wrapfigure}[10]{r}{0.50\linewidth}
\centering
\vspace{-1.0em}
\includegraphics[width=\linewidth]{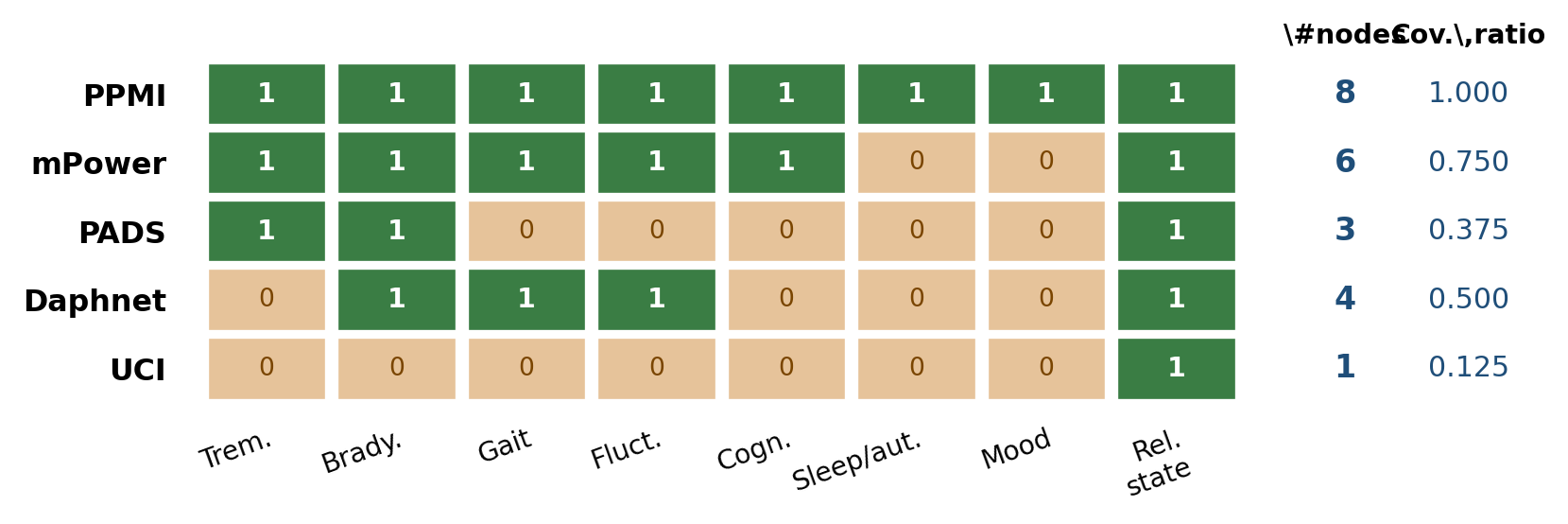}
\caption{Coverage of the 8-dim symptom-node space across 5 datasets. Green = covered (1), tan = missing (0). Right columns: \#nodes covered and ratio.}
\label{fig:coverage}
\vspace{-1.0em}
\end{wrapfigure}
Fig.~\ref{fig:coverage} and Table~\ref{tab:coverage-matrix} further explain why SGC-RML can connect five heterogeneous datasets. The five datasets exhibit a high degree of imbalance in modality coverage: PPMI covers all 8 nodes (7 clinical symptom nodes + 1 reliability\_state auxiliary node), making it the most complete data source for clinical anchors; mPower covers 6/8 nodes, primarily from mobile tasks; PADS covers 3/8 nodes, focusing on bilateral wrist movement-related symptoms; Daphnet covers 4/8 nodes, primarily reflecting gait and freeze-related signals; and UCI only covers the reliability\_state node across the dataset. Despite the significant differences in coverage, the five datasets are still mapped to the same eight-node symptom space.

\vspace{-0.3em}
\begin{table}[H]
\centering
\caption{Per-node coverage across 5 datasets. Cells = observed sample counts
in each dataset's native unit (visit/subject/window); ``--'' = node not
covered. Bottom row = \#datasets observing each node (unit-free).}
\label{tab:coverage-matrix}
\scriptsize
\setlength{\tabcolsep}{2.5pt}
\resizebox{\textwidth}{!}{%
\begin{tabular}{l r r r r r r r r c r}
\toprule
\textbf{Dataset} & \textbf{Trem.} & \textbf{Brady.} & \textbf{Gait} & \textbf{Fluc.} & \textbf{Cog.} & \textbf{Sleep} & \textbf{Mood} & \textbf{Rel.} & \textbf{\#} & \textbf{Cov.} \\
\midrule
  PPMI & 26,322 & 26,322 & 12,502 & 35,311 & 18,741 & 20,315 & 20,577 & 46,172 & 8/8 & 100\% \\
  mPower & 961 & 969 & 387 & 387 & 124 & -- & -- & 970 & 6/8 & 75\% \\
  PADS & 419 & 419 & -- & -- & -- & -- & -- & 419 & 3/8 & 38\% \\
  Daphnet & -- & 29,906 & 29,906 & 29,906 & -- & -- & -- & 29,906 & 4/8 & 50\% \\
  UCI & -- & -- & -- & -- & -- & -- & -- & 5,875 & 1/8 & 12\% \\
\midrule
  \textit{\#datasets observing} & 3/5 & 4/5 & 3/5 & 3/5 & 2/5 & 1/5 & 1/5 & 5/5 & 8/8 & 100\% \\
\bottomrule
\end{tabular}%
}
\end{table}

Detailed discussion of why missingness-aware mapping outperforms na\"ive feature alignment, and the cross-dataset role of \textsc{reliability\_state}, is deferred to Appendix~\ref{app:complementarity-discussion}.

\subsection{PPMI ablation experiments show that the performance improvement comes from temporal symptom-node modelling}
\label{sec:exp-ppmi-ablation}
Table~\ref{tab:ppmi-ablation} presents the architectural ablation of PPMI's UPDRS-III predictions for the next 12 months. The results show that the simple LightGBM baseline achieves MAEs of 4.899 and 4.905 on raw clinical and clinical multimodal features, respectively; LightGBM using only 8-dimensional symptom nodes does not directly achieve the best results, with an MAE of 4.918. This is important because it illustrates that the performance improvement is not automatically brought about by "symptom node compression" itself, nor can it be achieved simply by changing the input representation. The real improvement comes from the combination of symptom node representation with temporal modelling and node attention mechanisms.

\begin{table}[!ht]
\centering
\caption{PPMI architecture ablation for 12-month future UPDRS-III
prediction. \textbf{Bold} = best.}
\label{tab:ppmi-ablation}
\small
\setlength{\tabcolsep}{4pt}
\resizebox{\textwidth}{!}{%
\begin{tabular}{l l r r r r r}
\toprule
\textbf{Model} & \textbf{Input features} & \textbf{MAE}$\downarrow$ & \textbf{RMSE}$\downarrow$ & \textbf{R\textsuperscript{2}}$\uparrow$ & \textbf{Pearson}$\uparrow$ & \textbf{Spearman}$\uparrow$ \\
\midrule
  LightGBM\_basic & raw\_clinical & 4.899 & 6.930 & 0.748 & 0.865 & 0.883 \\
  LightGBM\_clinical\_multimodal\_clean & clinical\_multimodal & 4.905 & 6.909 & 0.749 & 0.866 & 0.883 \\
  LightGBM\_symptom\_nodes & 8dim\_symptom\_nodes & 4.918 & 6.989 & 0.744 & 0.863 & 0.883 \\
  GraphTemporalTransformer~\cite{velickovic2018graph} & 8dim\_symptom\_nodes & 4.852 & 6.972 & 0.745 & 0.866 & 0.886 \\
  TemporalTransformer & 8dim\_symptom\_nodes & 4.586 & 6.606 & 0.772 & 0.879 & 0.897 \\
  \textbf{NodeAttentionTemporalTransformer} & 8dim\_symptom\_nodes & \textbf{4.579} & \textbf{6.600} & \textbf{0.772} & \textbf{0.880} & \textbf{0.897} \\
\bottomrule
\end{tabular}%
}
\end{table}

When the model was switched from static LightGBM to GraphTemporalTransformer, the MAE decreased to 4.852; further using TemporalTransformer, the MAE significantly decreased to 4.586, and the R² improved to 0.772; ultimately, NodeAttentionTemporalTransformer achieved the best MAE of 4.579, RMSE of 6.600, Pearson score of 0.880, and Spearman score of 0.897. These results indicate that predicting the longitudinal progression of PD requires modelling the dependent structure of symptom changes over time, rather than relying solely on cross-sectional feature combinations. While the gain from node attention is relatively limited numerically, its value lies not only in error reduction but also in providing an entry point for symptom-level interpretation, allowing model predictions to be examined through clinical semantics.

\subsection{Symptom-node attention is consistent with clinical priors}
\label{sec:exp-attention}
\begin{figure}[!ht]
\centering
\begin{subfigure}[t]{0.5\linewidth}
  \centering
  \includegraphics[width=\linewidth]{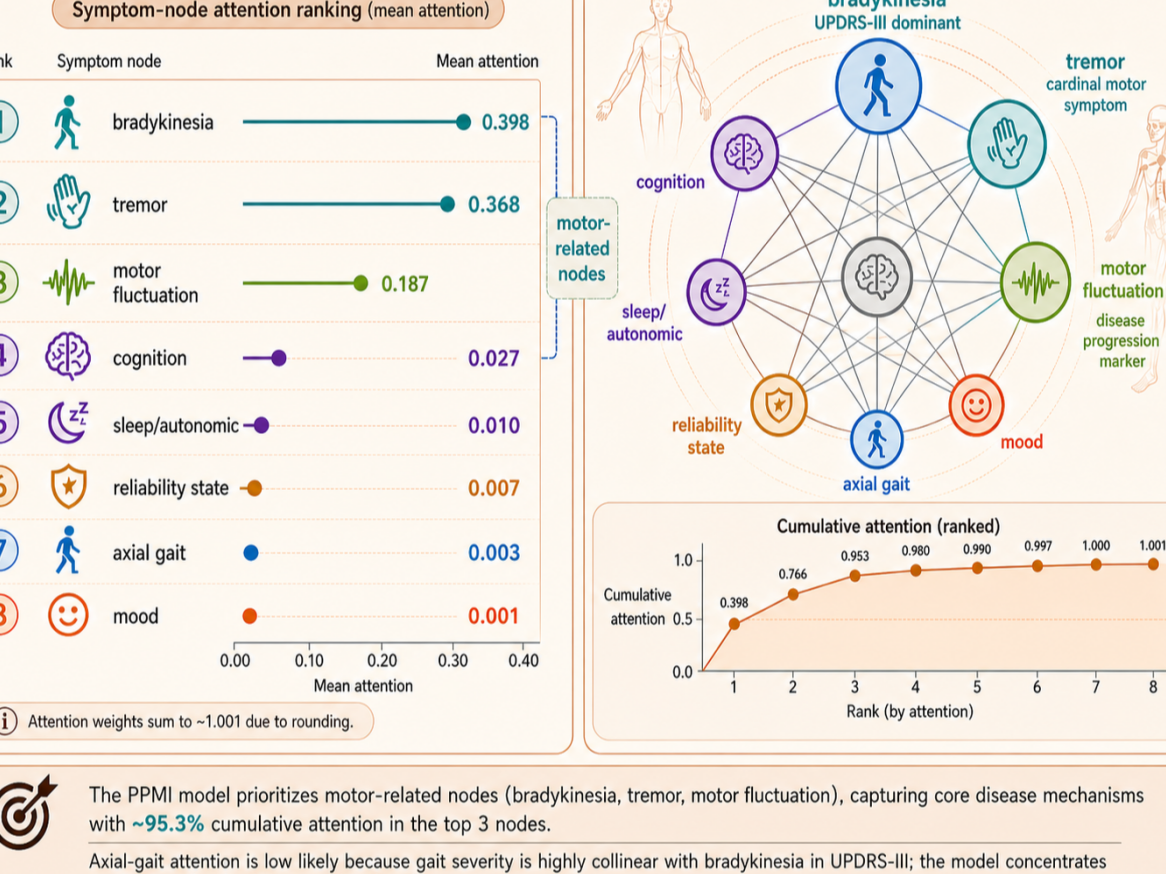}
  \caption{PPMI symptom-node attention (best-performing model).}
  \label{fig:attention}
\end{subfigure}\hfill
\begin{subfigure}[t]{0.5\linewidth}
  \centering
  \includegraphics[width=\linewidth]{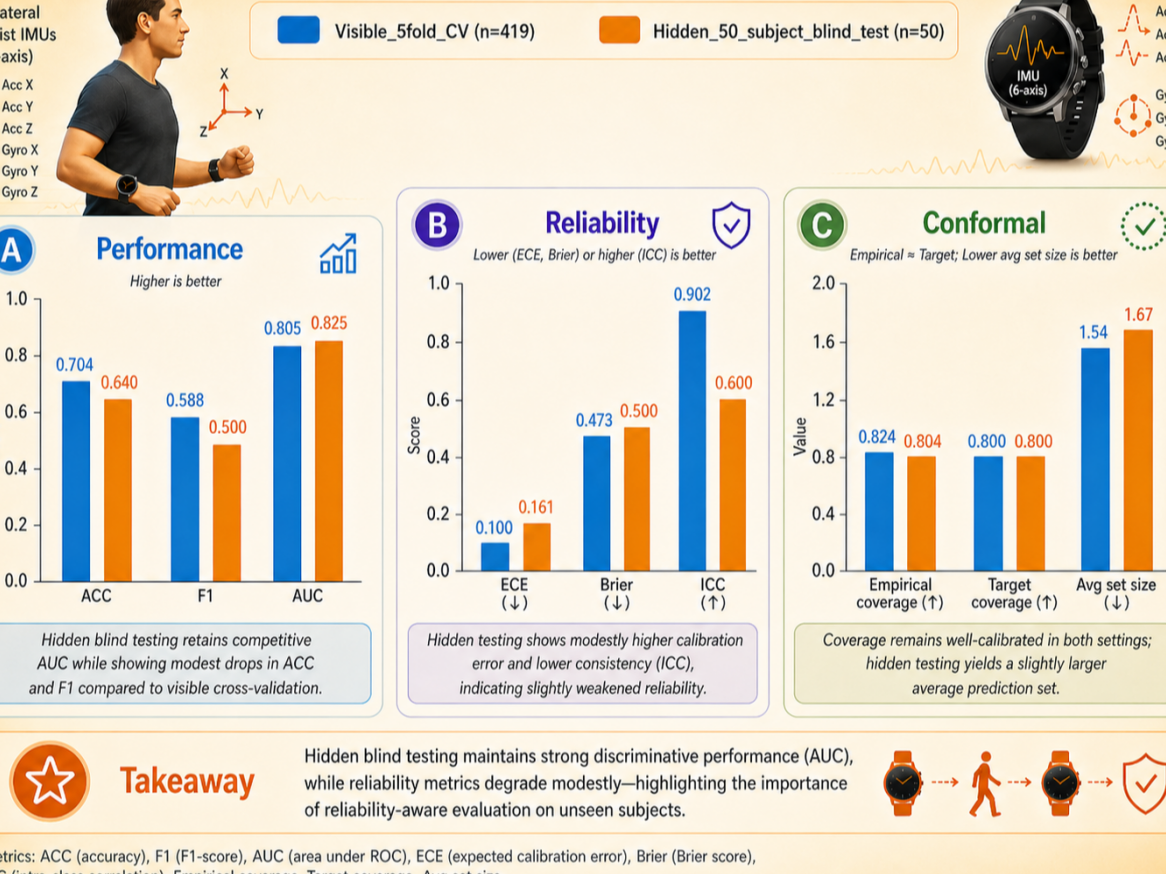}
  \caption{PADS visible vs.\ hidden-cohort reliability.}
  \label{fig:pads-reliability}
\end{subfigure}
\caption{\textbf{Trustworthiness evidence.}
(a) PPMI per-node attention; (b) PADS visible vs.\ hidden cohort.}
\label{fig:trustworthiness}
\end{figure}

As shown in Fig.~\ref{fig:attention}, attention concentrates on the motor nodes most relevant to UPDRS-III---bradykinesia, tremor, and motor fluctuation, jointly accounting for $\sim$95\% of cumulative weight. Non-motor nodes (cognition, sleep/autonomic, mood) and the auxiliary reliability\_state receive negligible weight, consistent with clinical priors. Full per-node values are provided in Appendix~\ref{app:reliability-interpretability} (Table~\ref{tab:ppmi-attention}). Note that attention weights provide explanatory hints rather than causal evidence~\cite{jain2019,wiegreffe2019}.

\subsection{The PADS hidden-blind test validates the reliability framework's generalization}
\label{sec:exp-pads-hidden}
Fig.~\ref{fig:pads-reliability} and Table~\ref{tab:pads-hidden} are among the most important pieces of reliability evidence in the entire paper. The PADS dataset not only reports the visible 5-fold cross-validation results but also includes a blind test cohort of 50 subjects with completely hidden participants. This hidden cohort did not participate in any training, hyperparameter tuning, or model selection, thus more closely resembling a real-world deployment scenario than standard cross-validation. The results show that the blind test AUC is 0.825, higher than the visible 5-fold AUC of 0.805. Although the ACC decreased from 0.704 to 0.640 and the F1 score decreased from 0.588 to 0.500, the Wilson 95\% confidence intervals of visible-CV and hidden-blind ACC overlap (visible $[0.659, 0.746]$ vs hidden $[0.501, 0.759]$, $\Delta = -0.064$), so the apparent ACC drop is \emph{not} statistically significant; the maintenance (or marginal improvement) of AUC confirms that the model retains stable ranking and discrimination capability on unseen subjects.

\begin{table}[!ht]
\centering
\caption{PADS visible 5-fold CV vs.\ hidden 50-subject blind test. ECE:
15-bin post temperature-scaling; Brier:
$\frac{1}{N}\sum_{i,c}(\hat{p}_{i,c}-y_{i,c})^2$;
ICC(2,1)~\cite{shrout1979}; Cov.: split-conformal coverage at target 0.80.
Square brackets give Wilson 95\% CIs for ACC and F1 (Hanley--McNeil for AUC); CIs of visible and hidden cohorts overlap, so the apparent ACC drop ($-0.064$) is not statistically significant at the 95\% level.}
\label{tab:pads-hidden}
\small
\setlength{\tabcolsep}{4pt}
\resizebox{\textwidth}{!}{%
\begin{tabular}{l r r r r r r r r}
\toprule
\textbf{Setting} & $n$ & ACC & F1 & \textbf{AUC} & ECE$\downarrow$ & Brier$\downarrow$ & ICC & \textbf{Cov.\@0.80} \\
\midrule
  Visible\_5fold\_CV & 419 & 0.704 [.659,.746] & 0.588 [.541,.633] & 0.805 [.764,.840] & 0.100 & 0.473 & 0.902 & 0.824 \\
  \textbf{Hidden\_50\_subject\_blind\_test} & 50 & 0.640 [.501,.759] & 0.500 [.366,.634] & \textbf{0.825 [.708,.942]} & 0.161 & 0.500 & 0.600 & \textbf{0.804 $\pm$ 0.053} \\
\bottomrule
\end{tabular}%
}
\end{table}

Calibration degradation across hidden subjects (ECE $+0.061$, Brier $+0.027$, ICC $-0.302$; Table~\ref{tab:pads-hidden}) confirms that out-of-distribution or absent subjects shift the calibration regime~\cite{ovadia2019can}, motivating the conformal-coverage and selective-rejection mechanisms. Yet conformal coverage remains near target (hidden $0.804 \pm 0.053$, visible 5-fold $0.824$, target $0.80$), so the model retains reliable rank ordering and supports the ``knowing when to be reliable'' claim even when probabilistic calibration is shaken. Detailed metric formulas and the per-mechanism interpretation are in Appendix~\ref{app:reliability-interpretability}.

\begin{figure}[!t]
\centering
\includegraphics[width=0.85\linewidth]{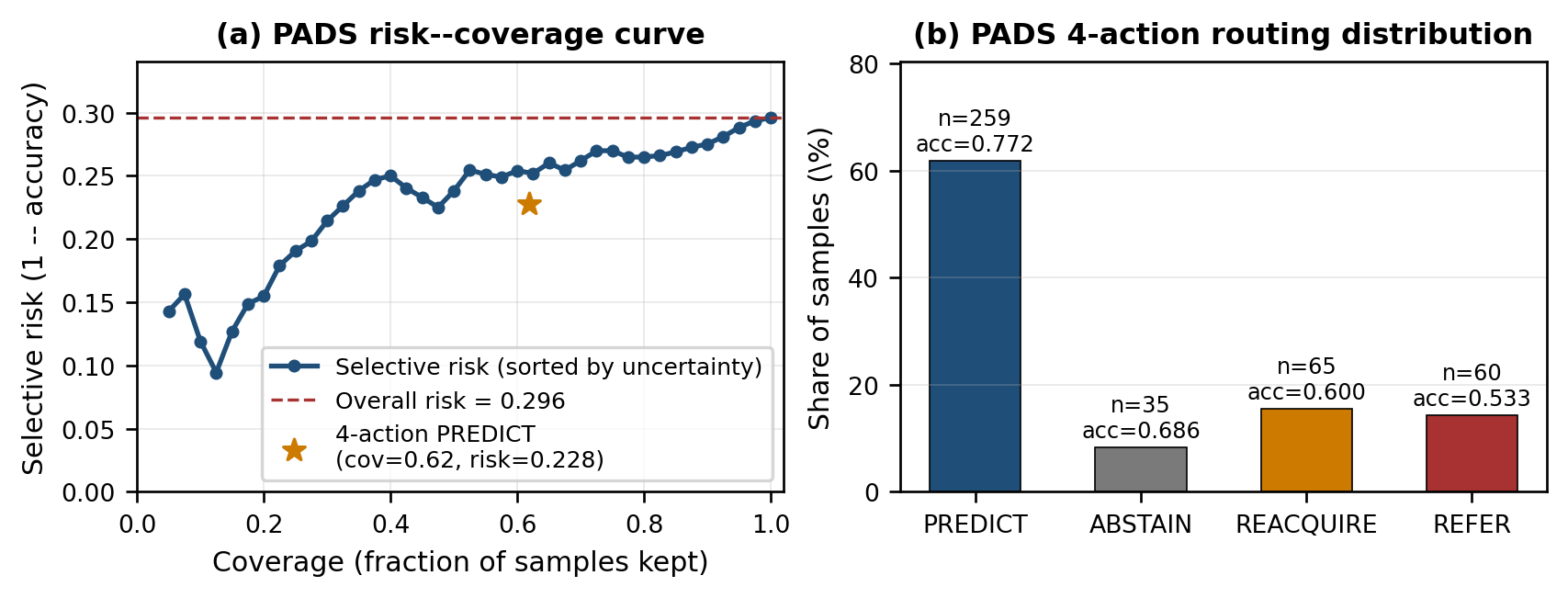}
\caption{\textbf{PADS reliability routing.} (a) Risk--coverage curve; orange star = 4-action router operating point (cov $0.618$, risk $0.228$); red dashed = no-rejection baseline. (b) 4-action distribution and per-subset accuracy: \textsc{predict} 61.8\% / 0.772, \textsc{abstain} 8.4\% / 0.686, \textsc{reacquire} 15.5\% / 0.600, \textsc{refer} 14.3\% / 0.533.}
\label{fig:pads-routing}
\end{figure}

Fig.~\ref{fig:pads-routing} operationalises this claim end-to-end. On the same 419-subject pool, accepting only the 61.8\% of samples routed to \textsc{predict} raises classification accuracy from the no-rejection baseline of $0.704$ to $\mathbf{0.772}$ ($+6.8$ percentage points), while the rejected partitions \textsc{abstain} ($0.686$), \textsc{reacquire} ($0.600$), and \textsc{refer} ($0.533$) carry progressively lower conditional accuracy in exactly the order the router predicted. This is the concrete behaviour SGC-RML's deterministic post-hoc policy is designed to produce: low-quality, high-uncertainty, or out-of-distribution inputs are deferred to data re-acquisition or human review, while the retained predictions become more accurate, not less. The risk--coverage curve in panel (a) further shows that the router operating point lies on the achievable selective-risk frontier rather than below it---there is no free accuracy gain from a smarter post-hoc selector at the same coverage level.

\subsection{UCI personalized anchors substantially improve longitudinal severity prediction}
\label{sec:exp-uci-personalized}
On UCI Telemonitoring, leak-free temporal anchoring reveals strong subject-specific baseline effects: for each subject, the earliest $n_{\mathrm{anchor}}$ visits form the anchor set and all subsequent visits constitute the test set, so no future-to-past information leakage occurs. Without anchors, the subject-independent setting is essentially non-predictive (motor MAE 8.38, CCC 0.02); 5 anchors already drop motor MAE to 3.40, and 20 anchors reach 3.24 / CCC 0.756 with split-conformal coverage held at the $0.80$ target. Importantly, simple label-only baselines (\textsc{last-anchor}, \textsc{anchor-mean}) under the same protocol also achieve substantial gains, indicating that anchor labels carry most of the predictive signal on UCI's voice-feature space; we therefore use UCI primarily as a leak-free anchoring and regression conformal-routing case study rather than as evidence that the SGC-RML voice encoder outperforms trivial baselines~\cite{sun2020testtime}. The full anchor sweep, baseline comparison, and triggered-retesting interpretation are provided in Appendix~\ref{app:dataset-details} and~\ref{app:reliability-interpretability}.

\subsection{Comprehensive discussion: the core contribution of SGC-RML lies in its reliable clinical auxiliary assessment paradigm}
\label{sec:exp-discussion}
Across the five datasets, SGC-RML's main contribution is an auditable reliability interface rather than a single-task performance gain: dataset-specific encoders are preserved, while symptom-node mapping, calibration, conformal coverage, and 4-action routing decide when predictions are sufficiently supported and when they should be abstained, reacquired, or referred. Full calibration / conformal coverage / 4-action routing are demonstrated on PADS and (in regression form) on UCI; PPMI, mPower, and Daphnet provide complementary evidence for longitudinal architecture choice, seed-level stability, and subject-level variance respectively (see the cross-dataset summary in Appendix~\ref{app:reliability-summary}).

\enlargethispage{8\baselineskip}
\subsection{Limitations and Future work}
\label{sec:exp-limitations}
While the experimental results support the validity of SGC-RML, limitations remain. First, different datasets differ significantly in labeling systems, sampling protocols, and task definitions; cross-cohort dataset shift~\cite{finlayson2021clinician} therefore prevents direct cross-dataset numerical comparison. Second, lacks prospective clinical validation; the heterogeneity of public or semi-public datasets is insufficient to fully represent real-world clinical workflows. Third, fairness and subgroup analysis are still inadequate, and PD digital assessment may be affected by factors (age, gender, disease stage, device type, medication status, recording environment, and physical ability). However, these limitations do not diminish the contributions of this paper but rather indicate that SGC-RML is geared towards challenging scenarios closer to real-world deployment. Future work will conduct larger-scale, multicenter, prospective cohort studies to assess generalization in real-world workflows and systematically incorporate fairness and subgroup analysis to clarify the reliability and validity across different populations, devices, and clinical conditions.

\FloatBarrier

%% file: checklist.tex
\section*{NeurIPS Paper Checklist}

\begin{enumerate}

\item {\bf Claims}
    \item[] Question: Do the main claims made in the abstract and introduction accurately reflect the paper's contributions and scope?
    \item[] Answer: \answerYes{} 
    \item[] Justification: The abstract and introduction state the contributions and scope of the paper.
    \item[] Guidelines:
    \begin{itemize}
        \item The answer \answerNA{} means that the abstract and introduction do not include the claims made in the paper.
        \item The abstract and/or introduction should clearly state the claims made, including the contributions made in the paper and important assumptions and limitations. A \answerNo{} or \answerNA{} answer to this question will not be perceived well by the reviewers. 
        \item The claims made should match theoretical and experimental results, and reflect how much the results can be expected to generalize to other settings. 
        \item It is fine to include aspirational goals as motivation as long as it is clear that these goals are not attained by the paper. 
    \end{itemize}

\item {\bf Limitations}
    \item[] Question: Does the paper discuss the limitations of the work performed by the authors?
    \item[] Answer: \answerYes{} 
    \item[] Justification: The paper discusses limitations and future work in Section~\ref{sec:exp-limitations}, including the scope of evaluation and practical constraints of the proposed framework.
    \item[] Guidelines:
    \begin{itemize}
        \item The answer \answerNA{} means that the paper has no limitation while the answer \answerNo{} means that the paper has limitations, but those are not discussed in the paper. 
        \item The authors are encouraged to create a separate ``Limitations'' section in their paper.
        \item The paper should point out any strong assumptions and how robust the results are to violations of these assumptions (e.g., independence assumptions, noiseless settings, model well-specification, asymptotic approximations only holding locally). The authors should reflect on how these assumptions might be violated in practice and what the implications would be.
        \item The authors should reflect on the scope of the claims made, e.g., if the approach was only tested on a few datasets or with a few runs. In general, empirical results often depend on implicit assumptions, which should be articulated.
        \item The authors should reflect on the factors that influence the performance of the approach. For example, a facial recognition algorithm may perform poorly when image resolution is low or images are taken in low lighting. Or a speech-to-text system might not be used reliably to provide closed captions for online lectures because it fails to handle technical jargon.
        \item The authors should discuss the computational efficiency of the proposed algorithms and how they scale with dataset size.
        \item If applicable, the authors should discuss possible limitations of their approach to address problems of privacy and fairness.
        \item While the authors might fear that complete honesty about limitations might be used by reviewers as grounds for rejection, a worse outcome might be that reviewers discover limitations that aren't acknowledged in the paper. The authors should use their best judgment and recognize that individual actions in favor of transparency play an important role in developing norms that preserve the integrity of the community. Reviewers will be specifically instructed to not penalize honesty concerning limitations.
    \end{itemize}

\item {\bf Theory assumptions and proofs}
    \item[] Question: For each theoretical result, does the paper provide the full set of assumptions and a complete (and correct) proof?
    \item[] Answer: \answerNA{} 
    \item[] Justification: The paper does not present theoretical results, theorems, or formal proofs. The contributions are evaluated empirically.
    \item[] Guidelines:
    \begin{itemize}
        \item The answer \answerNA{} means that the paper does not include theoretical results. 
        \item All the theorems, formulas, and proofs in the paper should be numbered and cross-referenced.
        \item All assumptions should be clearly stated or referenced in the statement of any theorems.
        \item The proofs can either appear in the main paper or the supplemental material, but if they appear in the supplemental material, the authors are encouraged to provide a short proof sketch to provide intuition. 
        \item Inversely, any informal proof provided in the core of the paper should be complemented by formal proofs provided in appendix or supplemental material.
        \item Theorems and Lemmas that the proof relies upon should be properly referenced. 
    \end{itemize}

    \item {\bf Experimental result reproducibility}
    \item[] Question: Does the paper fully disclose all the information needed to reproduce the main experimental results of the paper to the extent that it affects the main claims and/or conclusions of the paper (regardless of whether the code and data are provided or not)?
    \item[] Answer: \answerYes{} 
    \item[] Justification: An anonymized supplementary repository accompanies this submission, containing preprocessing scripts, training and evaluation scripts for all five datasets (PPMI, mPower, PADS, Daphnet FoG, UCI Telemonitoring), per-experiment configuration files, and a top-level README with reproduction instructions. The Reproducibility Card in Appendix~\ref{app:repro-card} summarizes per-dataset splits, optimizers, learning rates, batch sizes, epochs, threshold-selection rules, seeds, and approximate runtimes. For datasets that require application-based access (PPMI, mPower), the repository ships with scripts that assume local data paths after the user obtains access through the official channels cited in Section~\ref{sec:method}.
    \item[] Guidelines:
    \begin{itemize}
        \item The answer \answerNA{} means that the paper does not include experiments.
        \item If the paper includes experiments, a \answerNo{} answer to this question will not be perceived well by the reviewers: Making the paper reproducible is important, regardless of whether the code and data are provided or not.
        \item If the contribution is a dataset and\slash or model, the authors should describe the steps taken to make their results reproducible or verifiable. 
        \item Depending on the contribution, reproducibility can be accomplished in various ways. For example, if the contribution is a novel architecture, describing the architecture fully might suffice, or if the contribution is a specific model and empirical evaluation, it may be necessary to either make it possible for others to replicate the model with the same dataset, or provide access to the model. In general. releasing code and data is often one good way to accomplish this, but reproducibility can also be provided via detailed instructions for how to replicate the results, access to a hosted model (e.g., in the case of a large language model), releasing of a model checkpoint, or other means that are appropriate to the research performed.
        \item While NeurIPS does not require releasing code, the conference does require all submissions to provide some reasonable avenue for reproducibility, which may depend on the nature of the contribution. For example
        \begin{enumerate}
            \item If the contribution is primarily a new algorithm, the paper should make it clear how to reproduce that algorithm.
            \item If the contribution is primarily a new model architecture, the paper should describe the architecture clearly and fully.
            \item If the contribution is a new model (e.g., a large language model), then there should either be a way to access this model for reproducing the results or a way to reproduce the model (e.g., with an open-source dataset or instructions for how to construct the dataset).
            \item We recognize that reproducibility may be tricky in some cases, in which case authors are welcome to describe the particular way they provide for reproducibility. In the case of closed-source models, it may be that access to the model is limited in some way (e.g., to registered users), but it should be possible for other researchers to have some path to reproducing or verifying the results.
        \end{enumerate}
    \end{itemize}

\item {\bf Open access to data and code}
    \item[] Question: Does the paper provide open access to the data and code, with sufficient instructions to faithfully reproduce the main experimental results, as described in supplemental material?
    \item[] Answer: \answerYes{} 
    \item[] Justification: An anonymized supplementary repository is provided as supplementary material. It contains: (1) preprocessing scripts producing the unified 8-dim symptom-node tables for each dataset; (2) per-dataset training and evaluation scripts (\verb|pads/|, \verb|uci/|, \verb|mpower/|, \verb|ppmi/|, \verb|daphnet/|); (3) the shared reliability stack (conformal prediction, 4-action router) under \verb|core/|; (4) configuration files, an environment specification, and README instructions. The five datasets are publicly available through their official sources cited in Section~\ref{sec:method}; for restricted-access datasets (PPMI, mPower), scripts assume local paths after the user completes the access request. A non-anonymized version will be released upon acceptance.
    \item[] Guidelines:
    \begin{itemize}
        \item The answer \answerNA{} means that paper does not include experiments requiring code.
        \item Please see the NeurIPS code and data submission guidelines (\url{https://neurips.cc/public/guides/CodeSubmissionPolicy}) for more details.
        \item While we encourage the release of code and data, we understand that this might not be possible, so \answerNo{} is an acceptable answer. Papers cannot be rejected simply for not including code, unless this is central to the contribution (e.g., for a new open-source benchmark).
        \item The instructions should contain the exact command and environment needed to run to reproduce the results. See the NeurIPS code and data submission guidelines (\url{https://neurips.cc/public/guides/CodeSubmissionPolicy}) for more details.
        \item The authors should provide instructions on data access and preparation, including how to access the raw data, preprocessed data, intermediate data, and generated data, etc.
        \item The authors should provide scripts to reproduce all experimental results for the new proposed method and baselines. If only a subset of experiments are reproducible, they should state which ones are omitted from the script and why.
        \item At submission time, to preserve anonymity, the authors should release anonymized versions (if applicable).
        \item Providing as much information as possible in supplemental material (appended to the paper) is recommended, but including URLs to data and code is permitted.
    \end{itemize}

\item {\bf Experimental setting/details}
    \item[] Question: Does the paper specify all the training and test details (e.g., data splits, hyperparameters, how they were chosen, type of optimizer) necessary to understand the results?
    \item[] Answer: \answerYes{} 
    \item[] Justification: The paper reports the key experimental settings needed to understand the results, while full training and testing details will be provided in the released code, including data splits, hyperparameters, optimizers, and evaluation scripts.
    \item[] Guidelines:
    \begin{itemize}
        \item The answer \answerNA{} means that the paper does not include experiments.
        \item The experimental setting should be presented in the core of the paper to a level of detail that is necessary to appreciate the results and make sense of them.
        \item The full details can be provided either with the code, in appendix, or as supplemental material.
    \end{itemize}

\item {\bf Experiment statistical significance}
    \item[] Question: Does the paper report error bars suitably and correctly defined or other appropriate information about the statistical significance of the experiments?
    \item[] Answer: \answerYes{} 
    \item[] Justification: The paper reports statistical variability for the main experimental results in Section~\ref{sec:experiments}. Error bars are reported as mean and standard deviation over repeated runs or dataset splits, as specified in the experimental setup.
    \item[] Guidelines:
    \begin{itemize}
        \item The answer \answerNA{} means that the paper does not include experiments.
        \item The authors should answer \answerYes{} if the results are accompanied by error bars, confidence intervals, or statistical significance tests, at least for the experiments that support the main claims of the paper.
        \item The factors of variability that the error bars are capturing should be clearly stated (for example, train/test split, initialization, random drawing of some parameter, or overall run with given experimental conditions).
        \item The method for calculating the error bars should be explained (closed form formula, call to a library function, bootstrap, etc.)
        \item The assumptions made should be given (e.g., Normally distributed errors).
        \item It should be clear whether the error bar is the standard deviation or the standard error of the mean.
        \item It is OK to report 1-sigma error bars, but one should state it. The authors should preferably report a 2-sigma error bar than state that they have a 96\% CI, if the hypothesis of Normality of errors is not verified.
        \item For asymmetric distributions, the authors should be careful not to show in tables or figures symmetric error bars that would yield results that are out of range (e.g., negative error rates).
        \item If error bars are reported in tables or plots, the authors should explain in the text how they were calculated and reference the corresponding figures or tables in the text.
    \end{itemize}

\item {\bf Experiments compute resources}
    \item[] Question: For each experiment, does the paper provide sufficient information on the computer resources (type of compute workers, memory, time of execution) needed to reproduce the experiments?
    \item[] Answer: \answerYes{} 
    \item[] Justification:  We report the compute resources here. All experiments were run on a single NVIDIA A40 GPU with up to 48 GB memory on an internal shared cluster. The full five-dataset experiments, including ablations and the 5-seed mPower ensemble, required approximately 7 days of single-GPU wall-clock compute.
    \item[] Guidelines:
    \begin{itemize}
        \item The answer \answerNA{} means that the paper does not include experiments.
        \item The paper should indicate the type of compute workers CPU or GPU, internal cluster, or cloud provider, including relevant memory and storage.
        \item The paper should provide the amount of compute required for each of the individual experimental runs as well as estimate the total compute. 
        \item The paper should disclose whether the full research project required more compute than the experiments reported in the paper (e.g., preliminary or failed experiments that didn't make it into the paper). 
    \end{itemize}
    
\item {\bf Code of ethics}
    \item[] Question: Does the research conducted in the paper conform, in every respect, with the NeurIPS Code of Ethics \url{https://neurips.cc/public/EthicsGuidelines}?
    \item[] Answer: \answerYes{} 
    \item[] Justification: The research conforms to the NeurIPS Code of Ethics. All five datasets (PPMI, mPower, PADS, Daphnet FoG, UCI Telemonitoring) are public or access-controlled cohorts used under their permitted data-use terms; each was originally collected under its own IRB / equivalent approval framework (consistent with the IRB checklist item below). No new human-subject data is collected in this work; the submission is anonymized for double-blind review.
    \item[] Guidelines:
    \begin{itemize}
        \item The answer \answerNA{} means that the authors have not reviewed the NeurIPS Code of Ethics.
        \item If the authors answer \answerNo, they should explain the special circumstances that require a deviation from the Code of Ethics.
        \item The authors should make sure to preserve anonymity (e.g., if there is a special consideration due to laws or regulations in their jurisdiction).
    \end{itemize}

\item {\bf Broader impacts}
    \item[] Question: Does the paper discuss both potential positive societal impacts and negative societal impacts of the work performed?
    \item[] Answer: \answerYes{} 
    \item[] Justification: The paper discusses both positive and negative societal impacts. SGC-RML may improve access to PD assessment through lower-cost remote evaluation and safer abstain/reacquire/refer routing, while potential risks include privacy concerns, delayed care from incorrect decisions, and subgroup performance gaps. Fairness and subgroup auditing are discussed as future work in Section~\ref{sec:exp-limitations}.
    \item[] Guidelines:
    \begin{itemize}
        \item The answer \answerNA{} means that there is no societal impact of the work performed.
        \item If the authors answer \answerNA{} or \answerNo, they should explain why their work has no societal impact or why the paper does not address societal impact.
        \item Examples of negative societal impacts include potential malicious or unintended uses (e.g., disinformation, generating fake profiles, surveillance), fairness considerations (e.g., deployment of technologies that could make decisions that unfairly impact specific groups), privacy considerations, and security considerations.
        \item The conference expects that many papers will be foundational research and not tied to particular applications, let alone deployments. However, if there is a direct path to any negative applications, the authors should point it out. For example, it is legitimate to point out that an improvement in the quality of generative models could be used to generate Deepfakes for disinformation. On the other hand, it is not needed to point out that a generic algorithm for optimizing neural networks could enable people to train models that generate Deepfakes faster.
        \item The authors should consider possible harms that could arise when the technology is being used as intended and functioning correctly, harms that could arise when the technology is being used as intended but gives incorrect results, and harms following from (intentional or unintentional) misuse of the technology.
        \item If there are negative societal impacts, the authors could also discuss possible mitigation strategies (e.g., gated release of models, providing defenses in addition to attacks, mechanisms for monitoring misuse, mechanisms to monitor how a system learns from feedback over time, improving the efficiency and accessibility of ML).
    \end{itemize}
    
\item {\bf Safeguards}
    \item[] Question: Does the paper describe safeguards that have been put in place for responsible release of data or models that have a high risk for misuse (e.g., pre-trained language models, image generators, or scraped datasets)?
    \item[] Answer: \answerNA{} 
    \item[] Justification: The paper does not release high-risk models, generative systems, or scraped datasets with substantial misuse potential. Therefore, no additional safeguards for high-risk release are required.
    \item[] Guidelines:
    \begin{itemize}
        \item The answer \answerNA{} means that the paper poses no such risks.
        \item Released models that have a high risk for misuse or dual-use should be released with necessary safeguards to allow for controlled use of the model, for example by requiring that users adhere to usage guidelines or restrictions to access the model or implementing safety filters. 
        \item Datasets that have been scraped from the Internet could pose safety risks. The authors should describe how they avoided releasing unsafe images.
        \item We recognize that providing effective safeguards is challenging, and many papers do not require this, but we encourage authors to take this into account and make a best faith effort.
    \end{itemize}

\item {\bf Licenses for existing assets}
    \item[] Question: Are the creators or original owners of assets (e.g., code, data, models), used in the paper, properly credited and are the license and terms of use explicitly mentioned and properly respected?
    \item[] Answer: \answerYes{} 
    \item[] Justification: All existing datasets, models, and code assets used in the paper are properly cited. We follow their licenses and terms of use, and report the relevant asset information in the paper or supplementary material.
    \item[] Guidelines:
    \begin{itemize}
        \item The answer \answerNA{} means that the paper does not use existing assets.
        \item The authors should cite the original paper that produced the code package or dataset.
        \item The authors should state which version of the asset is used and, if possible, include a URL.
        \item The name of the license (e.g., CC-BY 4.0) should be included for each asset.
        \item For scraped data from a particular source (e.g., website), the copyright and terms of service of that source should be provided.
        \item If assets are released, the license, copyright information, and terms of use in the package should be provided. For popular datasets, \url{paperswithcode.com/datasets} has curated licenses for some datasets. Their licensing guide can help determine the license of a dataset.
        \item For existing datasets that are re-packaged, both the original license and the license of the derived asset (if it has changed) should be provided.
        \item If this information is not available online, the authors are encouraged to reach out to the asset's creators.
    \end{itemize}

\item {\bf New assets}
    \item[] Question: Are new assets introduced in the paper well documented and is the documentation provided alongside the assets?
    \item[] Answer: \answerYes{} 
    \item[] Justification: The paper documents the newly released code and experimental resources, including usage instructions, limitations, and reproduction scripts. We do not release new identifiable participant data.
    \item[] Guidelines:
    \begin{itemize}
        \item The answer \answerNA{} means that the paper does not release new assets.
        \item Researchers should communicate the details of the dataset\slash code\slash model as part of their submissions via structured templates. This includes details about training, license, limitations, etc. 
        \item The paper should discuss whether and how consent was obtained from people whose asset is used.
        \item At submission time, remember to anonymize your assets (if applicable). You can either create an anonymized URL or include an anonymized zip file.
    \end{itemize}

\item {\bf Crowdsourcing and research with human subjects}
    \item[] Question: For crowdsourcing experiments and research with human subjects, does the paper include the full text of instructions given to participants and screenshots, if applicable, as well as details about compensation (if any)? 
    \item[] Answer: \answerNA{} 
    \item[] Justification: The paper does not involve new human-subject experiments or new data collection. All experiments are conducted on existing public or access-controlled datasets under their original approvals and permitted data-use terms.
    \item[] Guidelines:
    \begin{itemize}
        \item The answer \answerNA{} means that the paper does not involve crowdsourcing nor research with human subjects.
        \item Including this information in the supplemental material is fine, but if the main contribution of the paper involves human subjects, then as much detail as possible should be included in the main paper. 
        \item According to the NeurIPS Code of Ethics, workers involved in data collection, curation, or other labor should be paid at least the minimum wage in the country of the data collector. 
    \end{itemize}

\item {\bf Institutional review board (IRB) approvals or equivalent for research with human subjects}
    \item[] Question: Does the paper describe potential risks incurred by study participants, whether such risks were disclosed to the subjects, and whether Institutional Review Board (IRB) approvals (or an equivalent approval/review based on the requirements of your country or institution) were obtained?
    \item[] Answer: \answerNA{} 
    \item[] Justification: The paper does not involve new human-subject experiments or new data collection. All experiments use existing public PD datasets that were each collected under their own IRB / institutional ethics approval framework: PPMI is a multi-center IRB-approved longitudinal cohort; mPower was collected under Sage Bionetworks' study protocol with informed electronic consent; PADS, Daphnet FoG, and UCI Telemonitoring are publicly released under their original collection ethics. We use only the publicly distributed, de-identified versions; this is consistent with the Code of Ethics item above.
    \item[] Guidelines:
    \begin{itemize}
        \item The answer \answerNA{} means that the paper does not involve crowdsourcing nor research with human subjects.
        \item Depending on the country in which research is conducted, IRB approval (or equivalent) may be required for any human subjects research. If you obtained IRB approval, you should clearly state this in the paper. 
        \item We recognize that the procedures for this may vary significantly between institutions and locations, and we expect authors to adhere to the NeurIPS Code of Ethics and the guidelines for their institution. 
        \item For initial submissions, do not include any information that would break anonymity (if applicable), such as the institution conducting the review.
    \end{itemize}

\item {\bf Declaration of LLM usage}
    \item[] Question: Does the paper describe the usage of LLMs if it is an important, original, or non-standard component of the core methods in this research? Note that if the LLM is used only for writing, editing, or formatting purposes and does \emph{not} impact the core methodology, scientific rigor, or originality of the research, declaration is not required.
    \item[] Answer: \answerNA{} 
    \item[] Justification: LLMs are not used as an important, original, or non-standard component of the core method. Their use is limited to writing, editing, and formatting assistance and does not affect the scientific content.
    \item[] Guidelines:
    \begin{itemize}
        \item The answer \answerNA{} means that the core method development in this research does not involve LLMs as any important, original, or non-standard components.
        \item Please refer to our LLM policy in the NeurIPS handbook for what should or should not be described.
    \end{itemize}

\end{enumerate}